\documentclass{bmvc2k}

\usepackage{graphicx}
\usepackage[title]{appendix}
\usepackage{amsmath,amssymb} 

\usepackage{multirow}

\usepackage{xcolor}
\usepackage{lipsum}
\usepackage{pdfpages}
\usepackage{float}
\usepackage{enumitem}

\makeatletter
\def\BState{\State\hskip-\ALG@thistlm}
\makeatother

\definecolor{orng}{rgb}{1,0.6275,0}
\definecolor{prpl}{rgb}{0.5882,0,0.6667}

\newcommand{\cfbox}[2]{%
    \colorlet{currentcolor}{.}%
    {\color{#1}%
    \fbox{\color{currentcolor}#2}}%
}

\title{Action Completion:\\A Temporal Model for Moment Detection}

\addauthor{Farnoosh Heidarivincheh}{Farnoosh.Heidarivincheh@bristol.ac.uk}{1}
\addauthor{Majid Mirmehdi}{M.Mirmehdi@bristol.ac.uk}{1}
\addauthor{Dima Damen}{Dima.Damen@bristol.ac.uk}{1}

\addinstitution{
 Department of Computer Science\\
 University of Bristol\\
 Bristol, UK
}

\runninghead{Heidarivincheh et al}{Action Completion: A Temporal Model..}

\def\etal{et al\bmvaOneDot}

\begin{document}

\maketitle

\begin{abstract}
We introduce \textit{completion moment} detection for actions - the problem of locating the \textit{moment} of completion, when the action's goal is confidently considered achieved.
The paper proposes a joint classification-regression recurrent model {that predicts} completion from a given frame, and then {integrates} frame-level contributions to detect sequence-level \textit{completion moment}. 
We introduce a recurrent voting node that 
predicts the frame's relative position of the \textit{completion moment} by either classification or regression.
The method is also capable of detecting incompletion. 
For example, the method is capable of detecting a missed ball-catch, as well as the \textit{moment} at which the ball is safely caught.
We test the method on 16 actions from three public datasets, covering sports as well as daily actions. Results show that when combining contributions from frames prior to the \textit{completion moment} as well as frames post completion, the \textit{completion moment} is detected within one second in 89\% of all tested sequences.
\end{abstract}

\vspace*{-2pt}
\section{Introduction}
\label{sec:intro}
\vspace*{-8pt}
An action, based on the Oxford Dictionary, is \textit{the fact or process of doing something, typically to achieve an aim}. Previous works on action recognition from visual data, {such as}~\cite{Simonyan2014,Ji2013,Donahue2015}, have overlooked assessing whether the action's \textit{aim} has actually been achieved, rather than merely attempted. The closely related action localisation problem, {e.g. in}~\cite{Yeung2016,Gkioxari2015,Hoai2014,Tian2013}, predicts the temporal start and end of an action's attempt, without assessing whether the \textit{aim} has been achieved either.
The notion of assessing an action's completion was introduced in~\cite{Heidari}, with follow-up works~\cite{Becattini2017,Farha2018} that focus on measuring the action's progress under a linear assumption, or predicting the time till the next action.
In this work, we attempt to detect (or locate) the \textit{moment} in time when the action can indeed be considered completed.


{We} define the problem of \textit{completion moment} detection as detecting the frame that separates pre-completion from post-completion per sequence, when present. 
Note that the \textit{completion moment} is different from the typical `start'/`end' frames in action localisation. The former focuses on the action's goal, while the latter {separates the motion relevant to the action from other actions or background frames.} 
{For example, in action `drink', the start of the action for localisation tends to be when a glass is lifted for drinking, and the end is when it is placed down. Conversely, the \textit{completion moment} we are after, is when the person consumes part of the beverage, marking their goal of drinking being achieved.}
{The subtle nature of this \textit{completion moment} thus requires a framework that is capable of robust moment detection.} 

Moment detection, including action \textit{completion moment} detection, has potential applications in robot-human collaboration, health-care or assisted-living, where an agent can react to a human completing the goal or conversely, failing to complete the action. For example, switching the oven off, could trigger safety alarms.

In detecting the moment of completion, we take a supervised approach{, where for training sequences, the \textit{completion moment} is labeled when present (see Sec~\ref{sec:assumptions}). 
Our proposed method uses a Convolutional-Recurrent Neural Network (C-RNN), and outputs per-frame votes for the presence and relative position of the \textit{completion moment}. 
We then predict a sequence-level \textit{completion moment} by accumulating these frame-level contributions. 
{To showcase the generality of our method, 
we evaluate it on 16 actions from 3 public datasets \cite{HMDB,UCF101,Heidari}. These include sports-based (e.g. \textit{basketball, pole vault}) as well as daily (e.g.~\textit{drink, pour}) actions. We 
show that both pre-completion and post-completion frames assist in \textit{completion moment} detection for the variety of tested actions.}


The remainder of this paper is organised as follows: related work in Sec.~\ref{sec:rel_work}, problem definition in Sec.~\ref{sec:assumptions}, proposed method in Sec.~\ref{sec:ac_model}, experiments and results in Sec.~\ref{sec:res} and conclusion and future work in Sec.~\ref{sec:Conc}.

\vspace*{-6pt}
\section{Related Work}
\label{sec:rel_work}
\vspace*{-4pt}
{Current methods for action recognition focus on deploying convolutional neural networks (CNNs), either dual-stream convolutions~\cite{Simonyan2014,Feichtenhofer2017,Wang2017} or 3D convolution filters from video snippets~\cite{Ji2013,Tran2015,Shou2016}, as well as recurrent neural networks (RNNs) that accumulate evidence from frames over a sequence~\cite{Yeung2018,Donahue2015,Ng2015}. 
However, these approaches aim to label the sequence as a whole. 
One seminal work~\cite{Wang2016} deviates by encoding the action as precondition and effect, using a Siamese network that predicts the action as a transformation between the two states.
In this section, we review related works that study partial observations within a video sequence for three problems of relevance to our proposed \textit{moment detection} problem,

\noindent {\textbf{Action Proposal Generation}: Action proposals and action-ness measures have 
{become the platform} for several action localisation approaches~\cite{Jain2014,Gkioxari2015,Yu2015,Zhao2017,Xiong2017}. Among these,~\cite{Zhao2017} and~\cite{Xiong2017} focus on classifying these proposals into those that contain the `completed' action, and incomplete proposals that should be rejected. While \cite{Xiong2017} applies an SVM to filter and reject spatio-temporal proposals containing incomplete or partial actions, \cite{Zhao2017} has embedded the rejection within an end-to-end CNN. 
These approaches classify each proposal, and do not attempt to locate or assess the \textit{completion moment}.

\noindent \textbf{Action Anticipation}: A few recent works \cite{Mahmud2017,Farha2018} focus on predicting the class label of the next unobserved action.
Mahmud \etal \cite{Mahmud2017} predict the next action as well as its starting time using a hybrid Siamese network in which an LSTM is used for temporal modelling. Farha \etal~\cite{Farha2018} 
estimate the time remaining until the next action, as well as the length and the label of the next action. The paper compares the usage of either an RNN or a CNN that takes as input concatenated frame-level features into a single tensor. {These approaches do not discuss completion (or incompletion) of the observed action.}

\noindent {\textbf{Early Detection}:
Several works~\cite{Hoai2014,Aliakbarian2017,Ma2016,Becattini2017,Yeung2016,Li2016} address early detection of partially observed actions, from as few frames as possible. 
These mainly propose loss functions to encourage early detection~\cite{Aliakbarian2017,Ma2016}, but a few works attempt fine-grained understanding of the action's progression.
{In~\cite{Hoai2014}, an SVM classifier is trained to accumulate scores from partial observations of the action, where the score is highest when the action is fully observed. The approach has been tested on facial and gesture datasets.}
Similarly, in \cite{Becattini2017}, an RNN is trained to predict the action label, as well as its linear progress towards {its} conclusion 
as a percentage (e.g. 50\% of the action has taken place).

{Two approaches~\cite{Yeung2016,Li2016} which detect moments within the sequence have been proposed, albeit for early detection and localisation.}
In~\cite{Yeung2016} individual frames predict the location of the next frame to be observed, using an RNN. The work aims for action detection with as few frames as possible, thus {the trained model proposes transitions within the sequence, by predicting the 
relative position of the frame to be observed next. 
Our work is inspired by ideas in~\cite{Li2016}, where action detection uses}
a joint classification-regression RNN. The classification branch predicts the ongoing action label which is then used by a regression branch to predict 
the start and the end points of the action, relative to the current frame. A Gaussian scoring function is used to encode the prediction uncertainty.  The approach was tested on 3D skeletal data for localisation, 
 {oblivious to} the action's completion (or incompletion).
None of the works mentioned above consider whether the action actually achieves its aim. In this work, we build on our previous work that introduced the action completion problem~\cite{Heidari} {by classifying whole sequences into complete and incomplete}, and take inspiration from~\cite{Li2016} to propose a joint classification-regression architecture. As opposed to predicting the next or the ongoing action, we detect the \textit{completion moment} by accumulating evidence from frame-level decisions. {We further define the \textit{completion moment} detection problem in the next section.} 

\begin{figure}[t]
\centering

\setlength{\fboxsep}{0pt}%
\setlength{\fboxrule}{1.5pt}%

\cfbox{orng}{\includegraphics[width=0.11\textwidth]{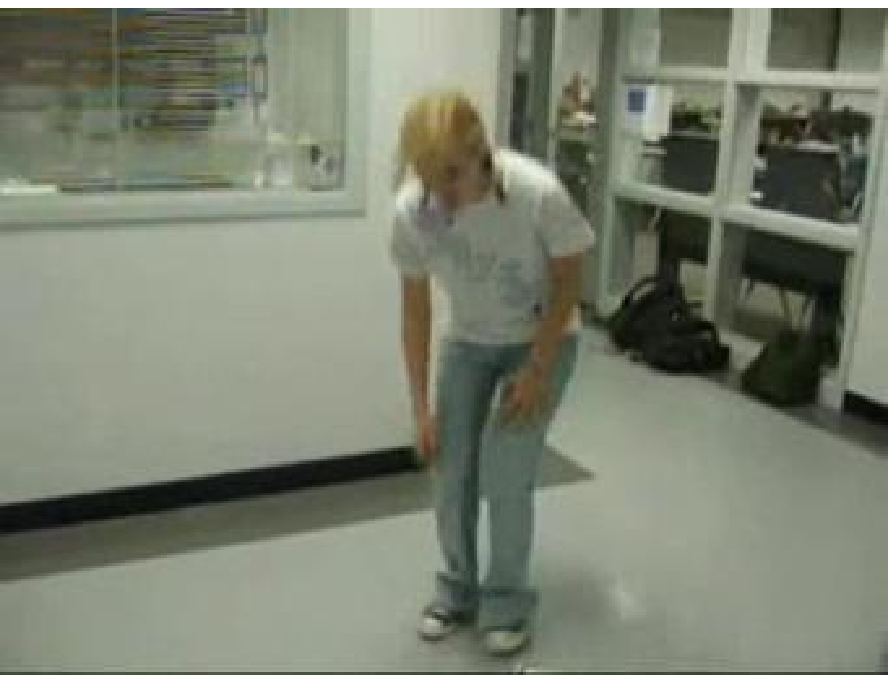}}\hspace{-2.2pt}
\cfbox{orng}{\includegraphics[width=0.11\textwidth]{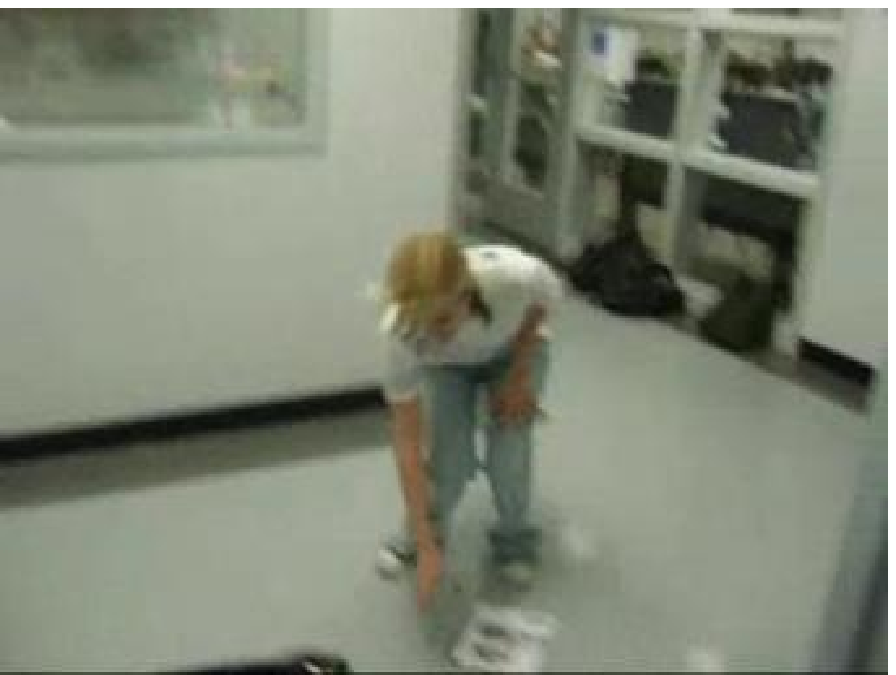}}\hspace{-2.2pt}
\cfbox{prpl}{\includegraphics[width=0.11\textwidth]{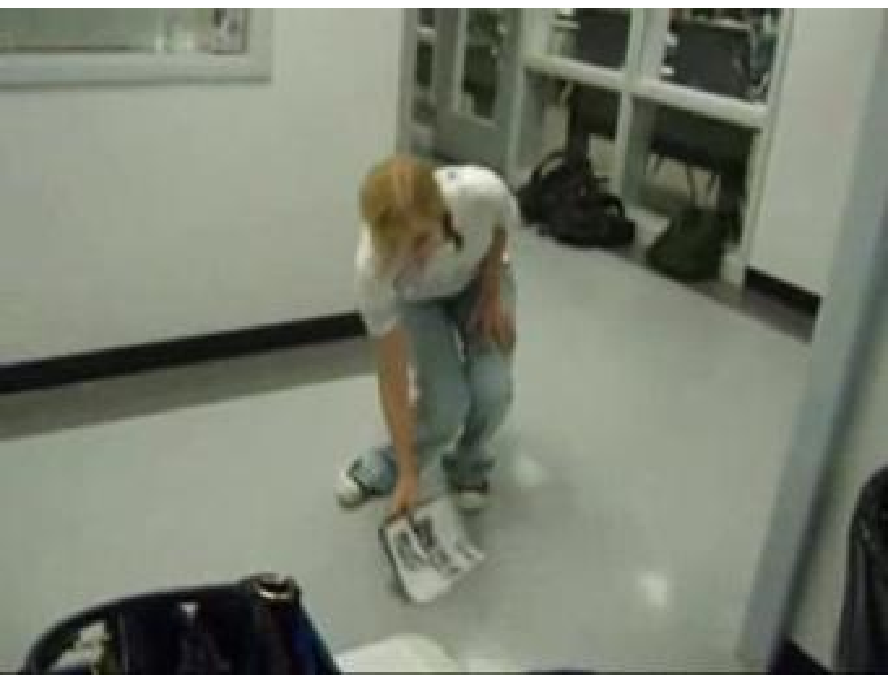}}\hspace{-2.2pt}
\cfbox{prpl}{\includegraphics[width=0.11\textwidth]{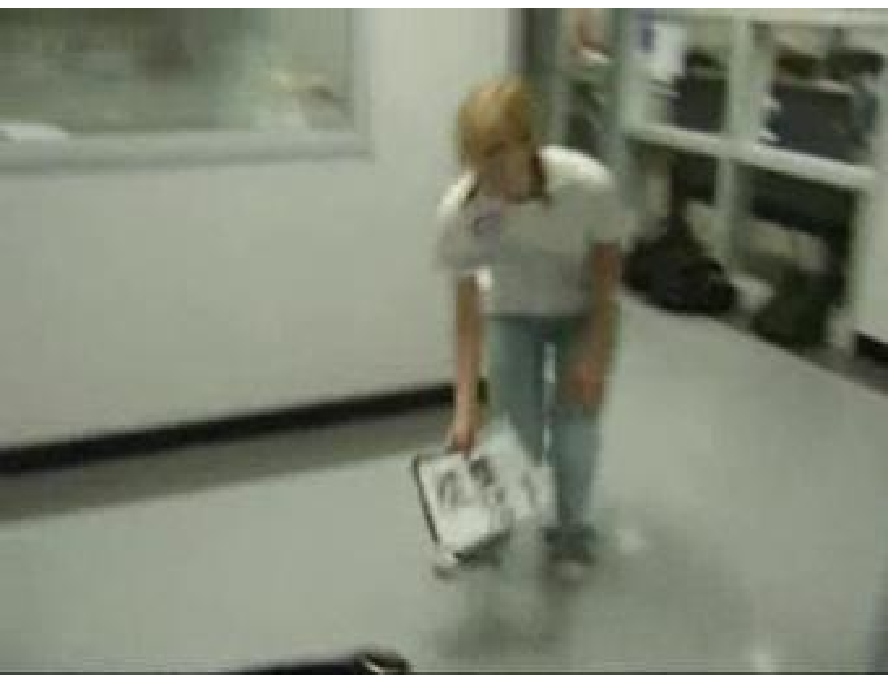}}\hspace{-2.2pt}
\hspace{0.03\textwidth}
\cfbox{orng}{\includegraphics[width=0.11\textwidth]{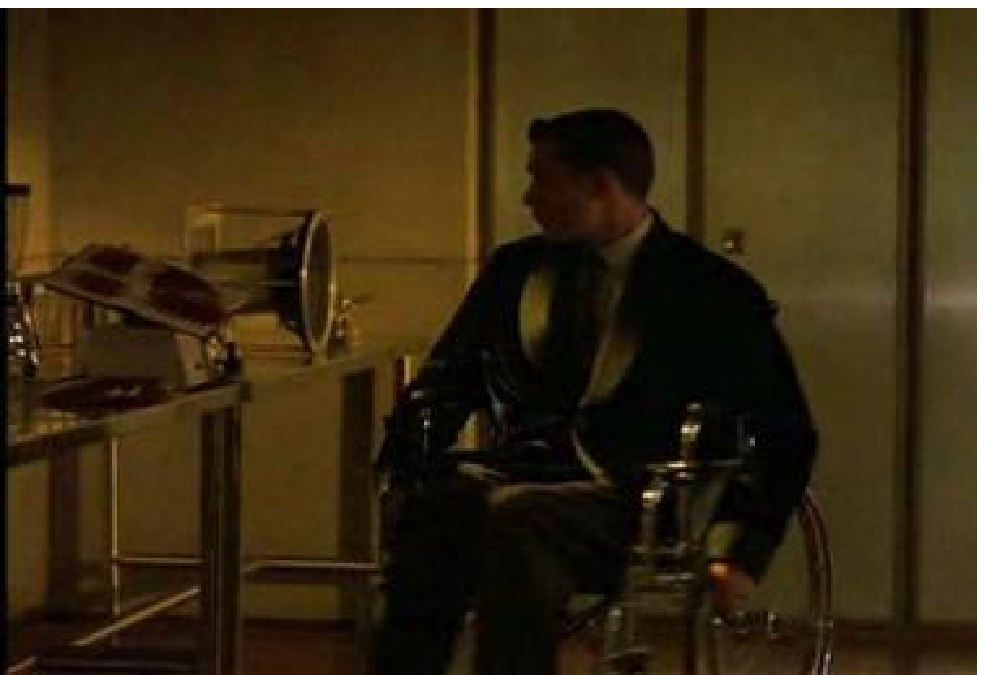}}\hspace{-2.2pt}
\cfbox{orng}{\includegraphics[width=0.11\textwidth]{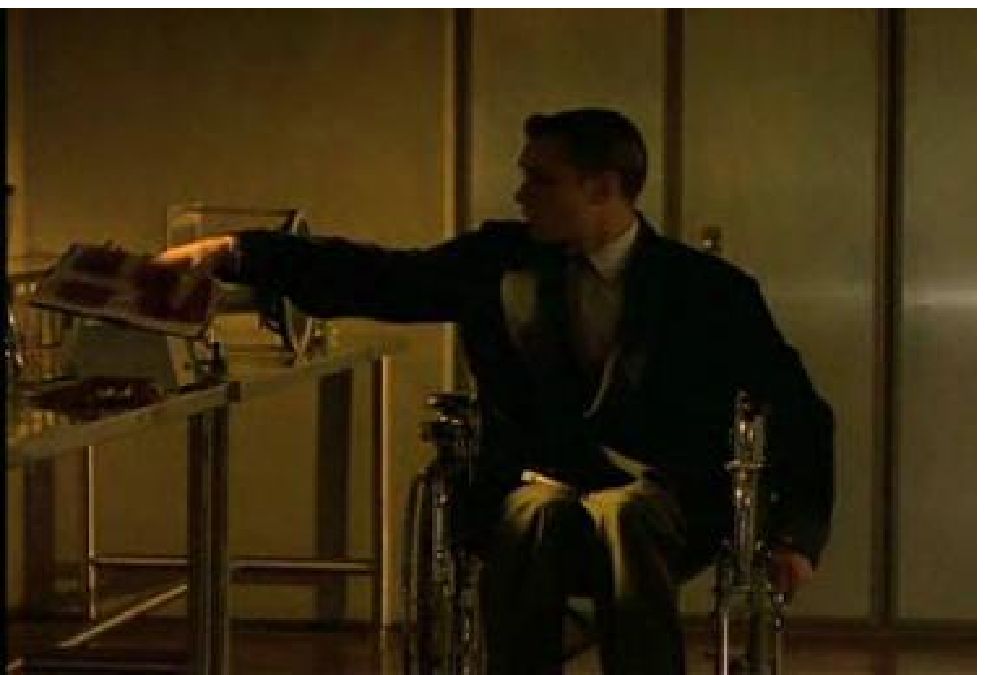}}\hspace{-2.2pt}
\cfbox{prpl}{\includegraphics[width=0.11\textwidth]{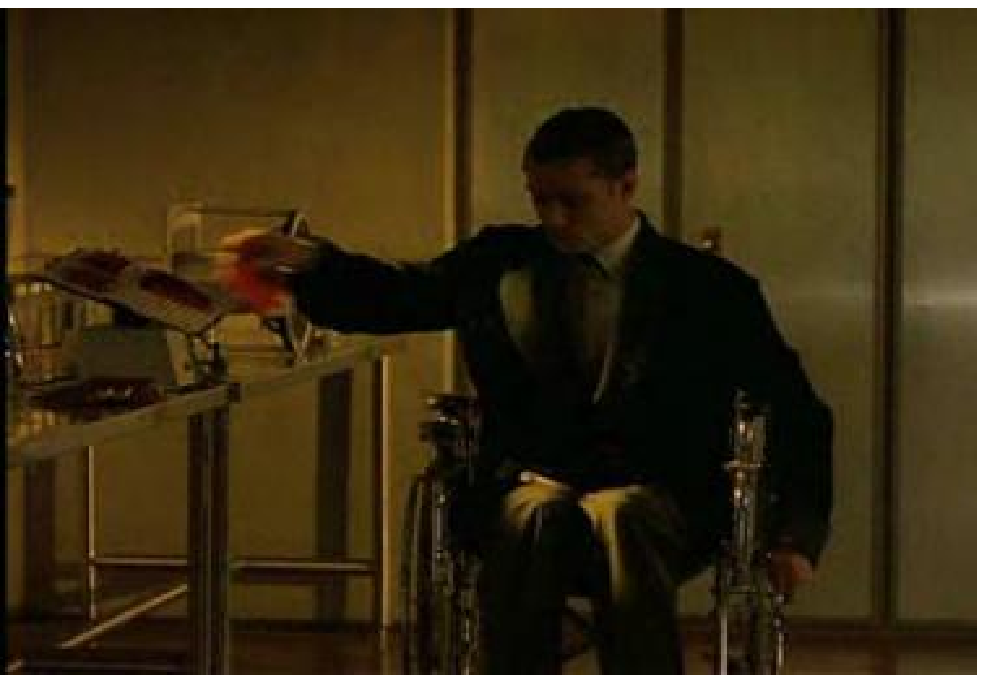}}\hspace{-2.2pt}
\cfbox{prpl}{\includegraphics[width=0.11\textwidth]{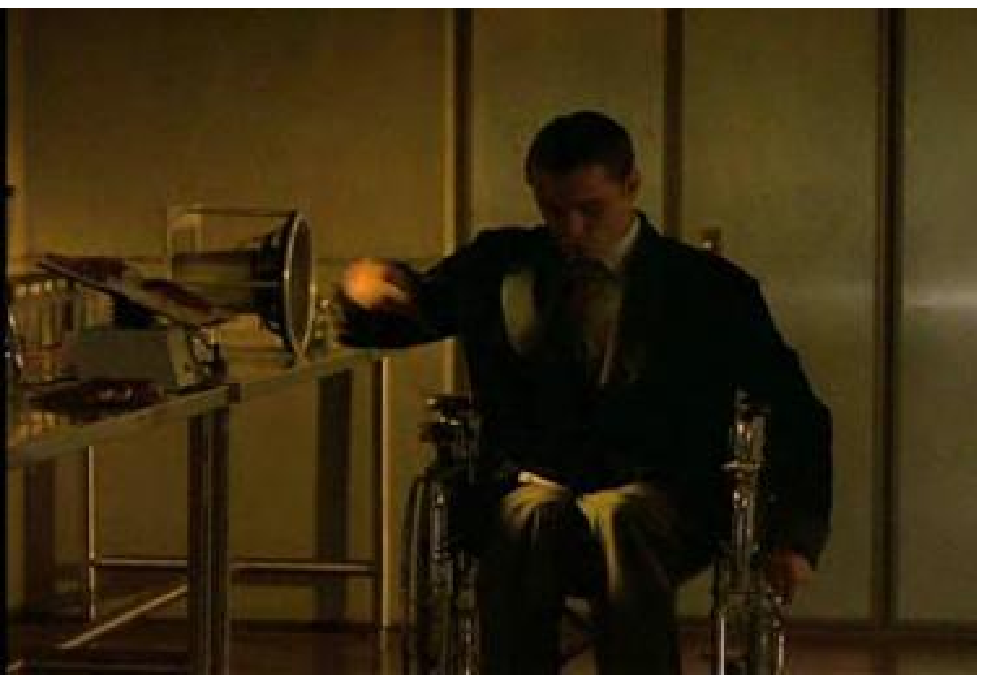}}\hspace{-2.2pt}
\vspace{1pt} 

\cfbox{orng}{\includegraphics[width=0.11\textwidth]{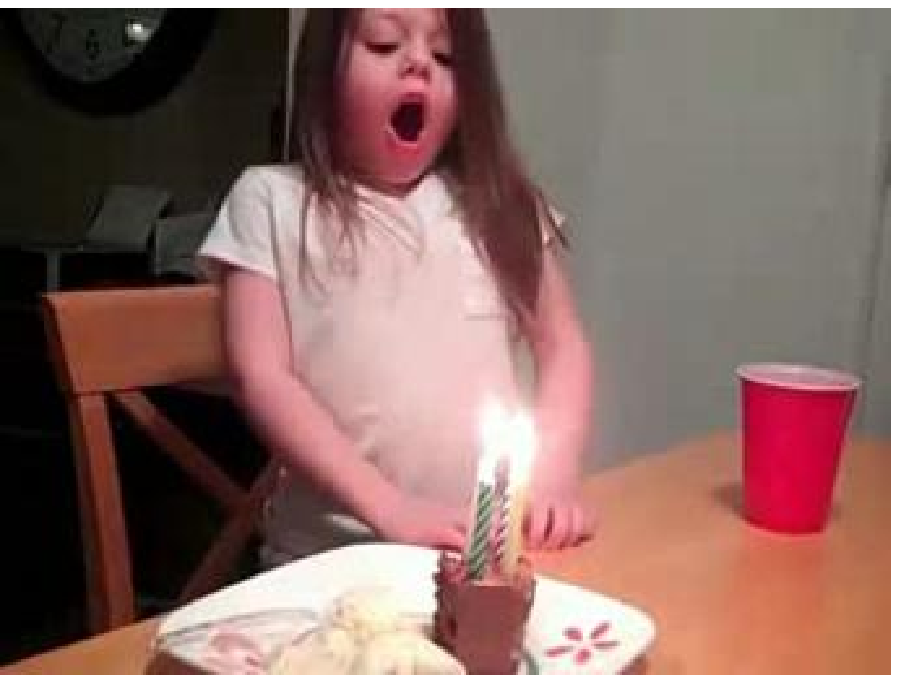}}\hspace{-2.2pt}
\cfbox{orng}{\includegraphics[width=0.11\textwidth]{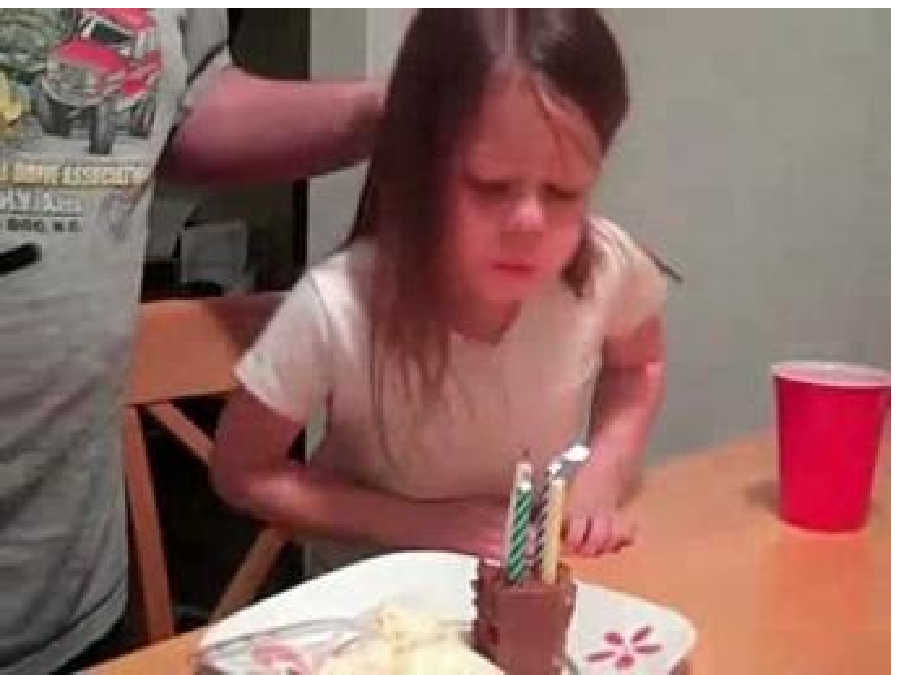}}\hspace{-2.2pt}
\cfbox{prpl}{\includegraphics[width=0.11\textwidth]{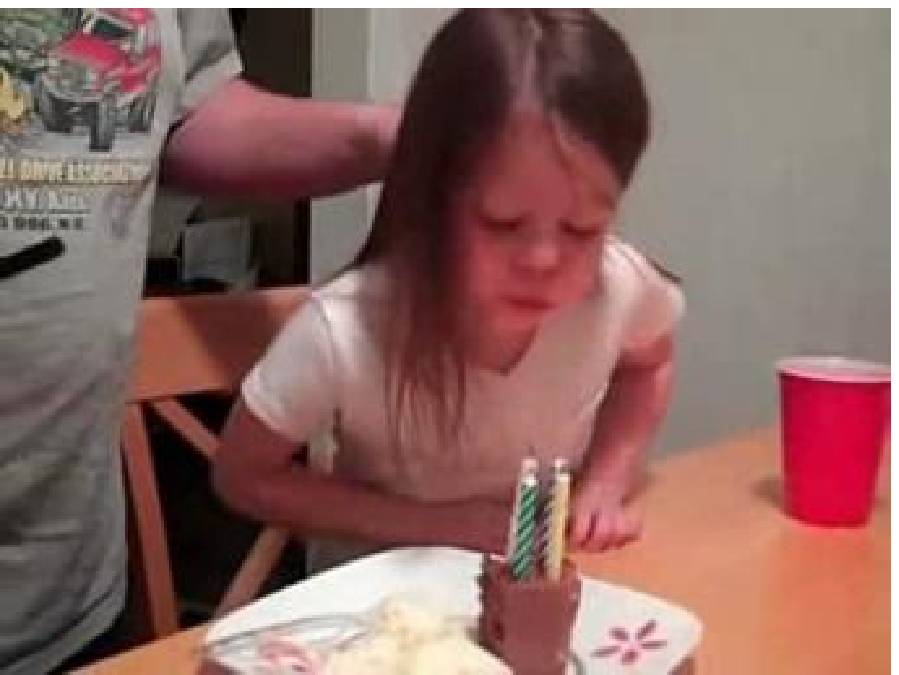}}\hspace{-2.2pt}
\cfbox{prpl}{\includegraphics[width=0.11\textwidth]{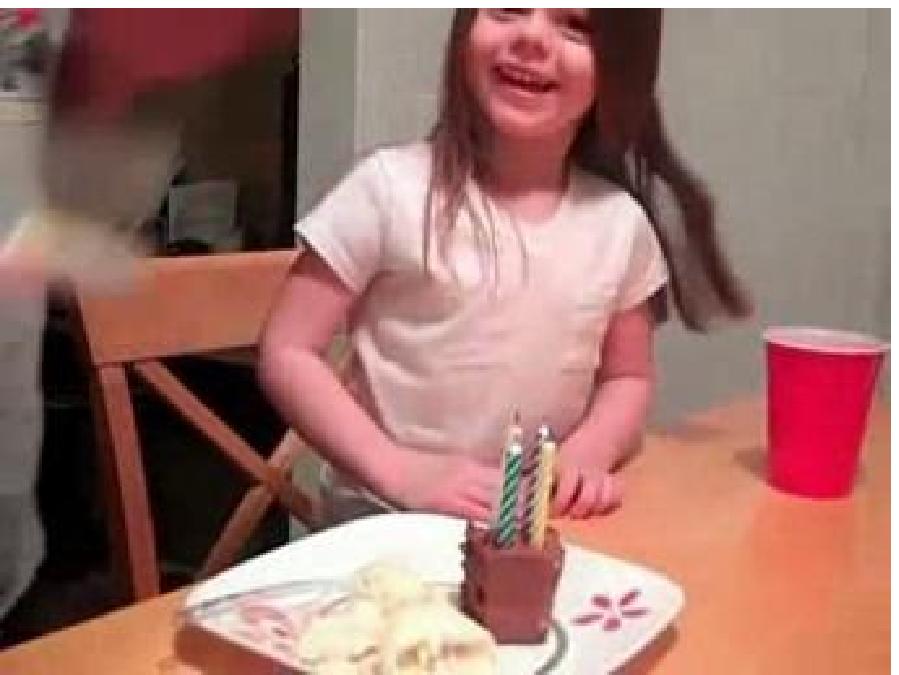}}
\hspace{0.03\textwidth}
\cfbox{orng}{\includegraphics[width=0.11\textwidth]{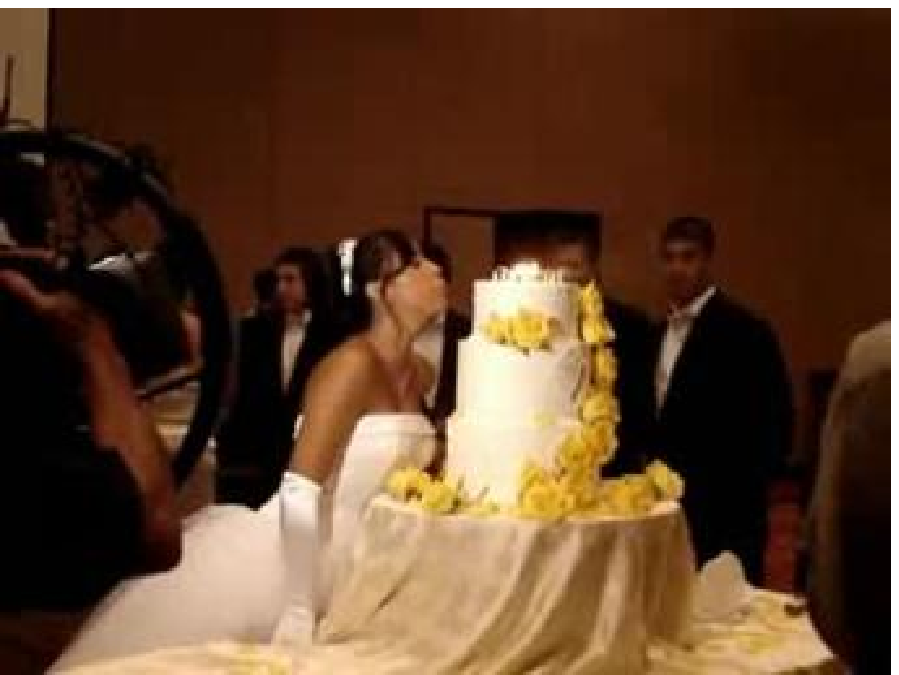}}\hspace{-2.2pt}
\cfbox{orng}{\includegraphics[width=0.11\textwidth]{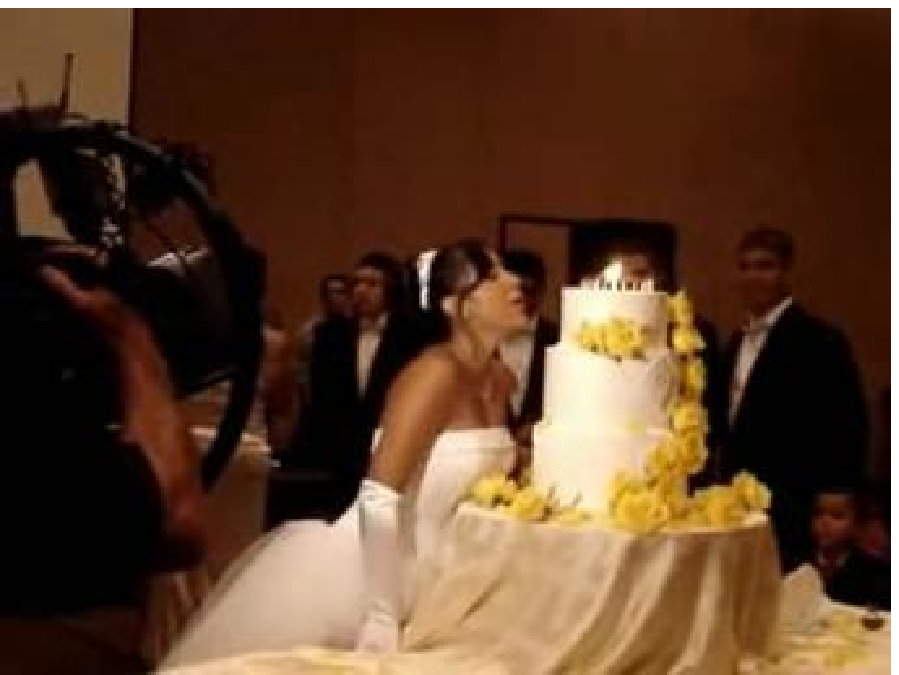}}\hspace{-2.2pt}
\cfbox{prpl}{\includegraphics[width=0.11\textwidth]{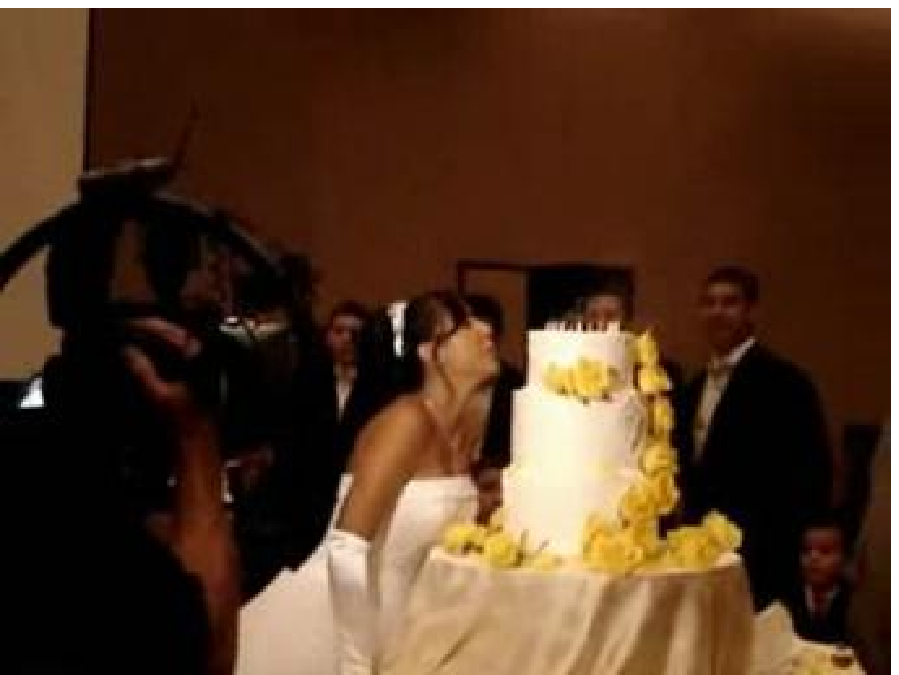}}\hspace{-2.2pt}
\cfbox{prpl}{\includegraphics[width=0.11\textwidth]{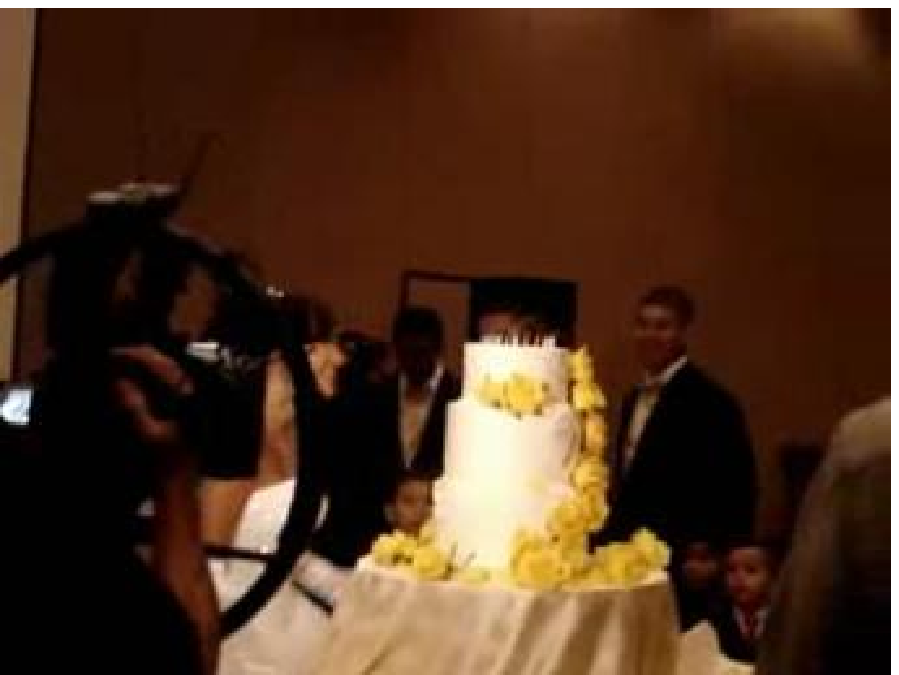}}\\
\vspace{1pt} 


\cfbox{orng}{\includegraphics[width=0.11\textwidth]{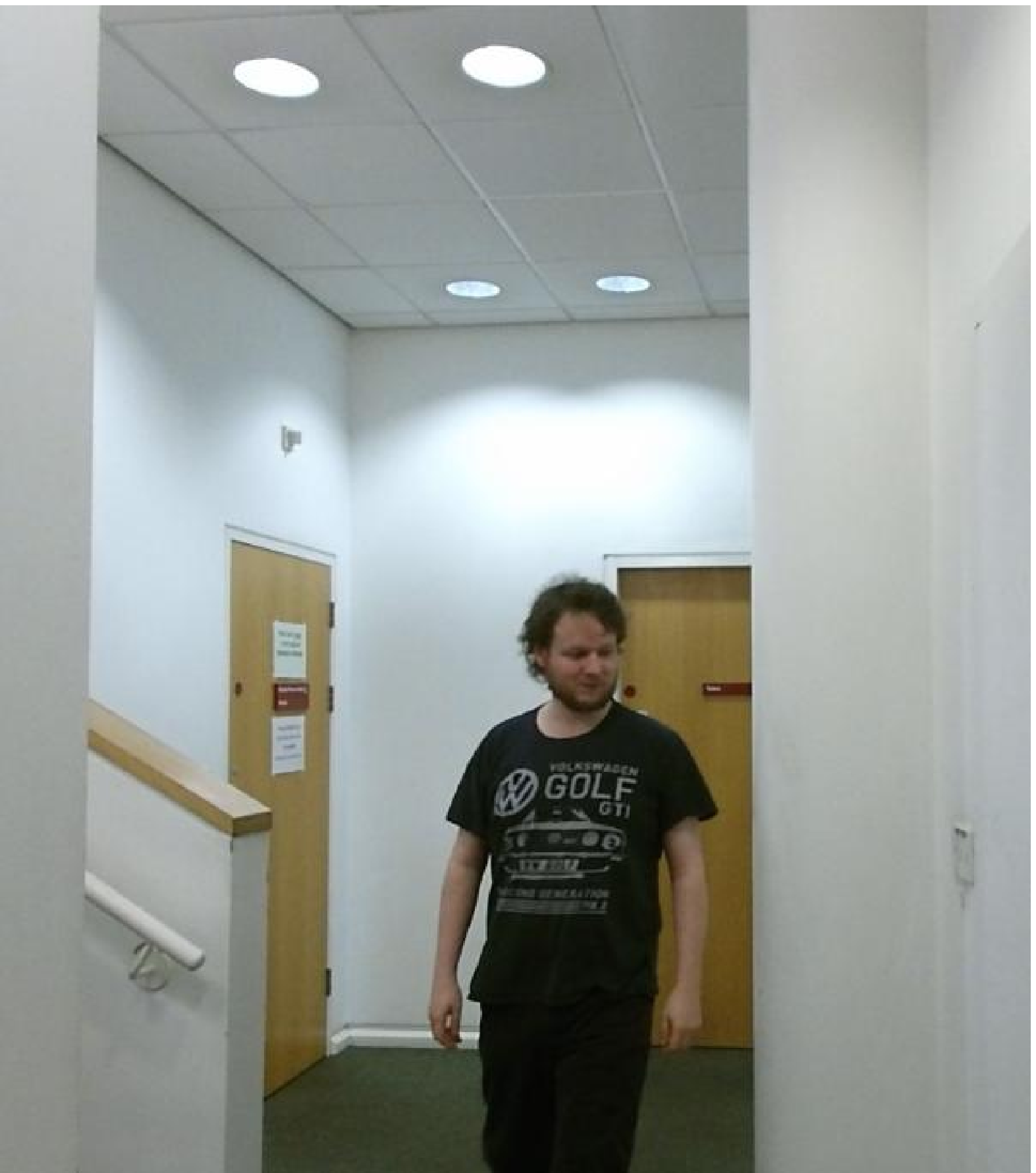}}\hspace{-2.2pt}
\cfbox{orng}{\includegraphics[width=0.11\textwidth]{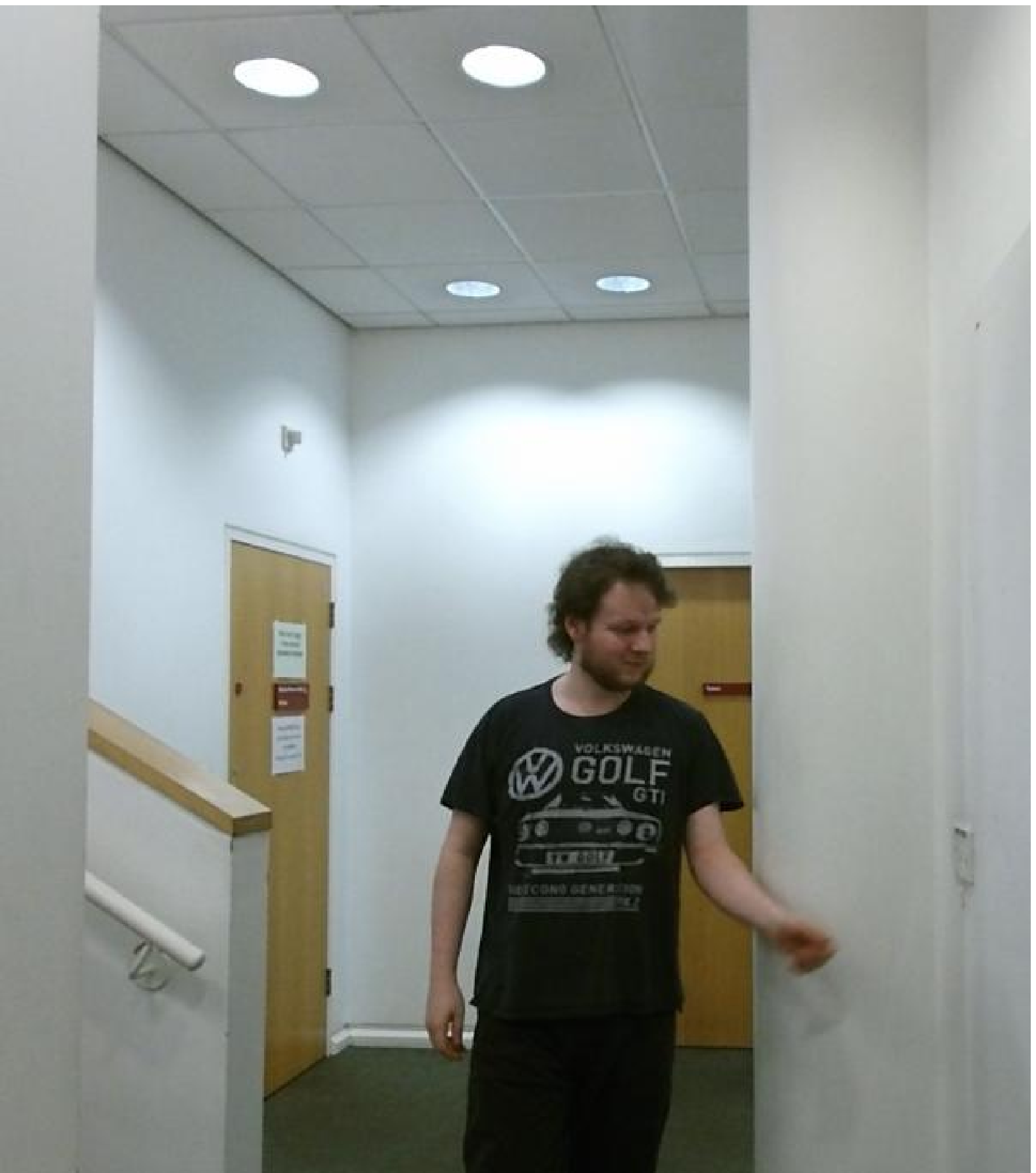}}\hspace{-2.2pt}
\cfbox{prpl}{\includegraphics[width=0.11\textwidth]{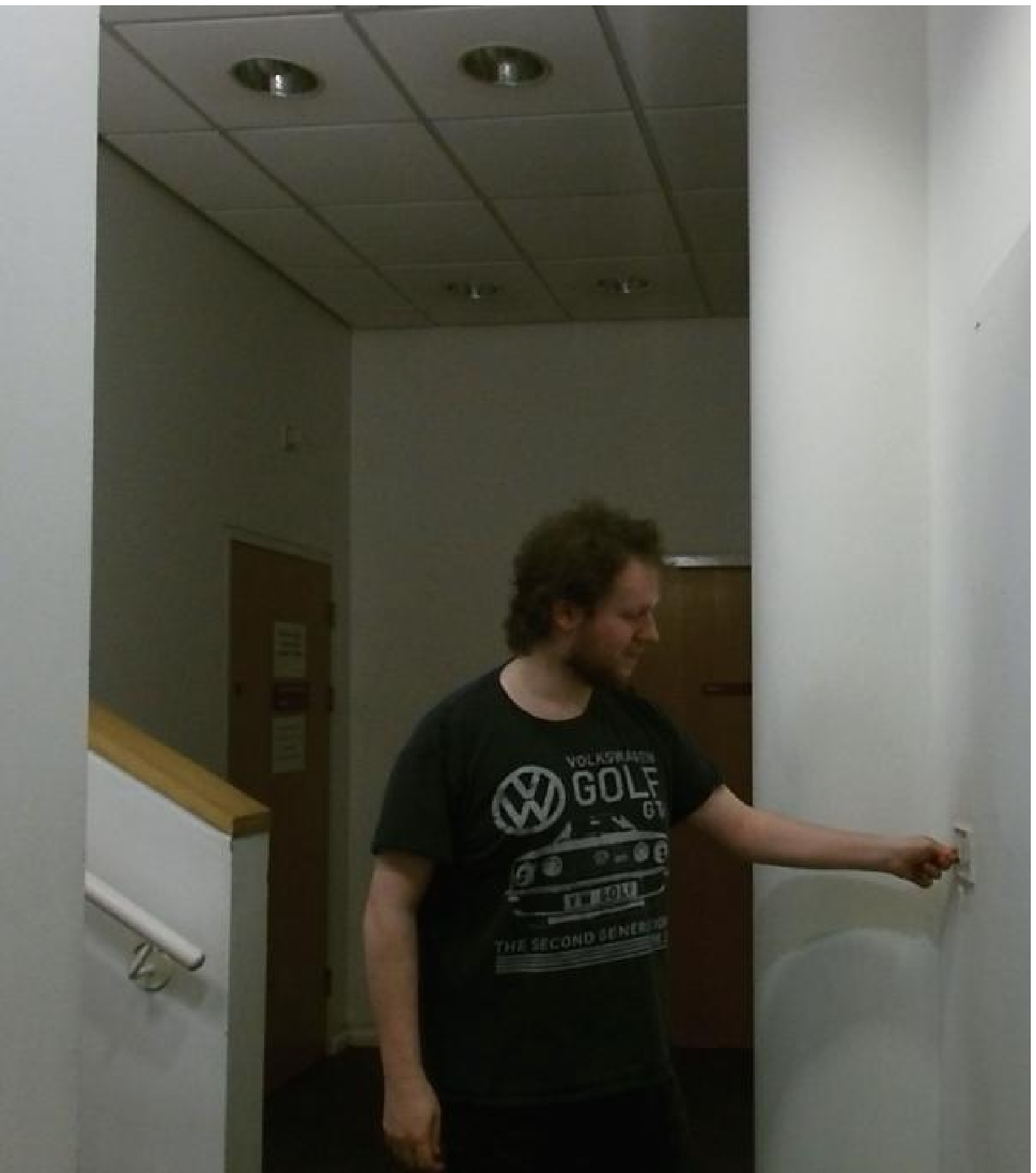}}\hspace{-2.2pt}
\cfbox{prpl}{\includegraphics[width=0.11\textwidth]{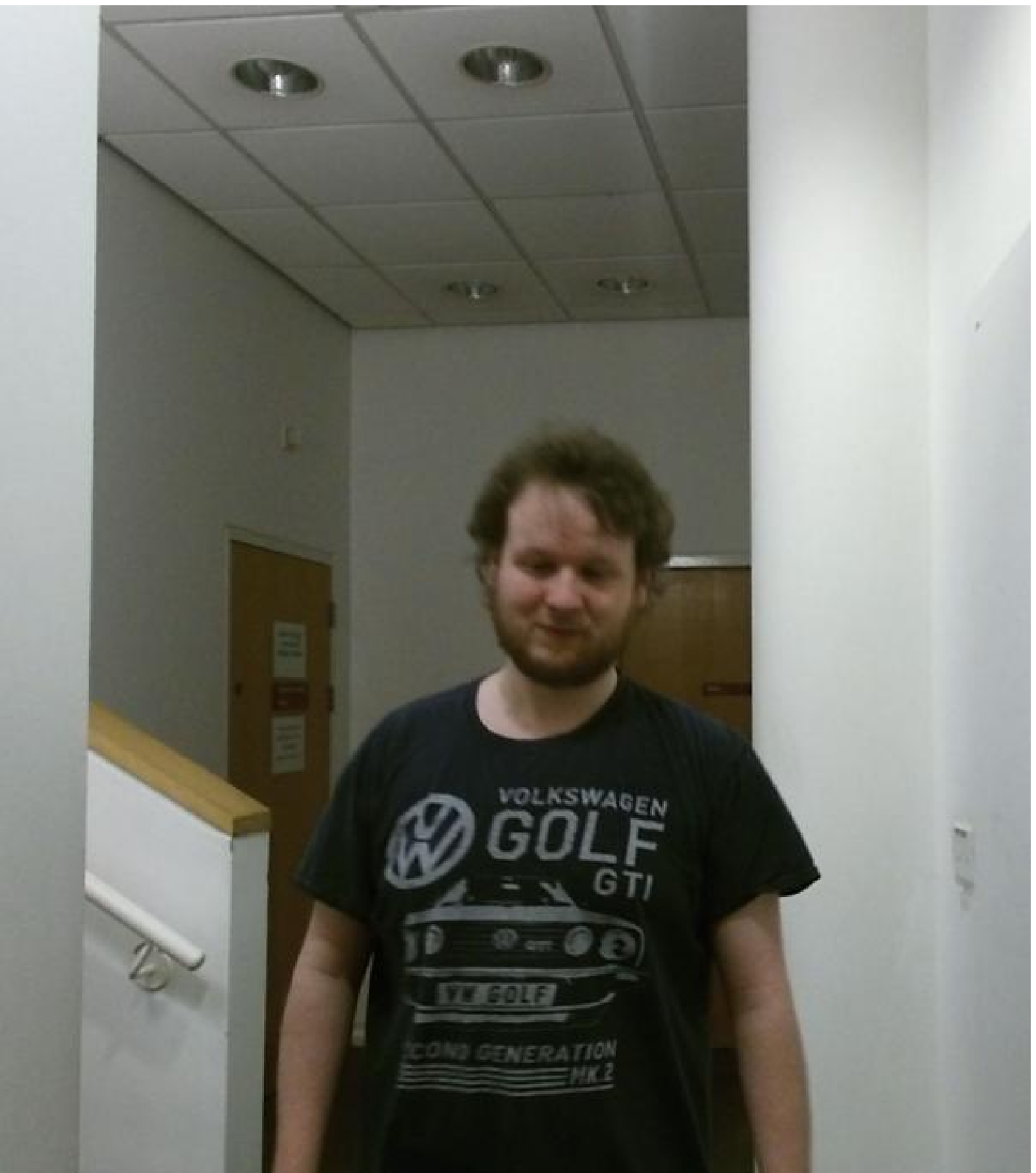}}\hspace{-2.2pt}
\hspace{0.03\textwidth}
\cfbox{orng}{\includegraphics[width=0.11\textwidth]{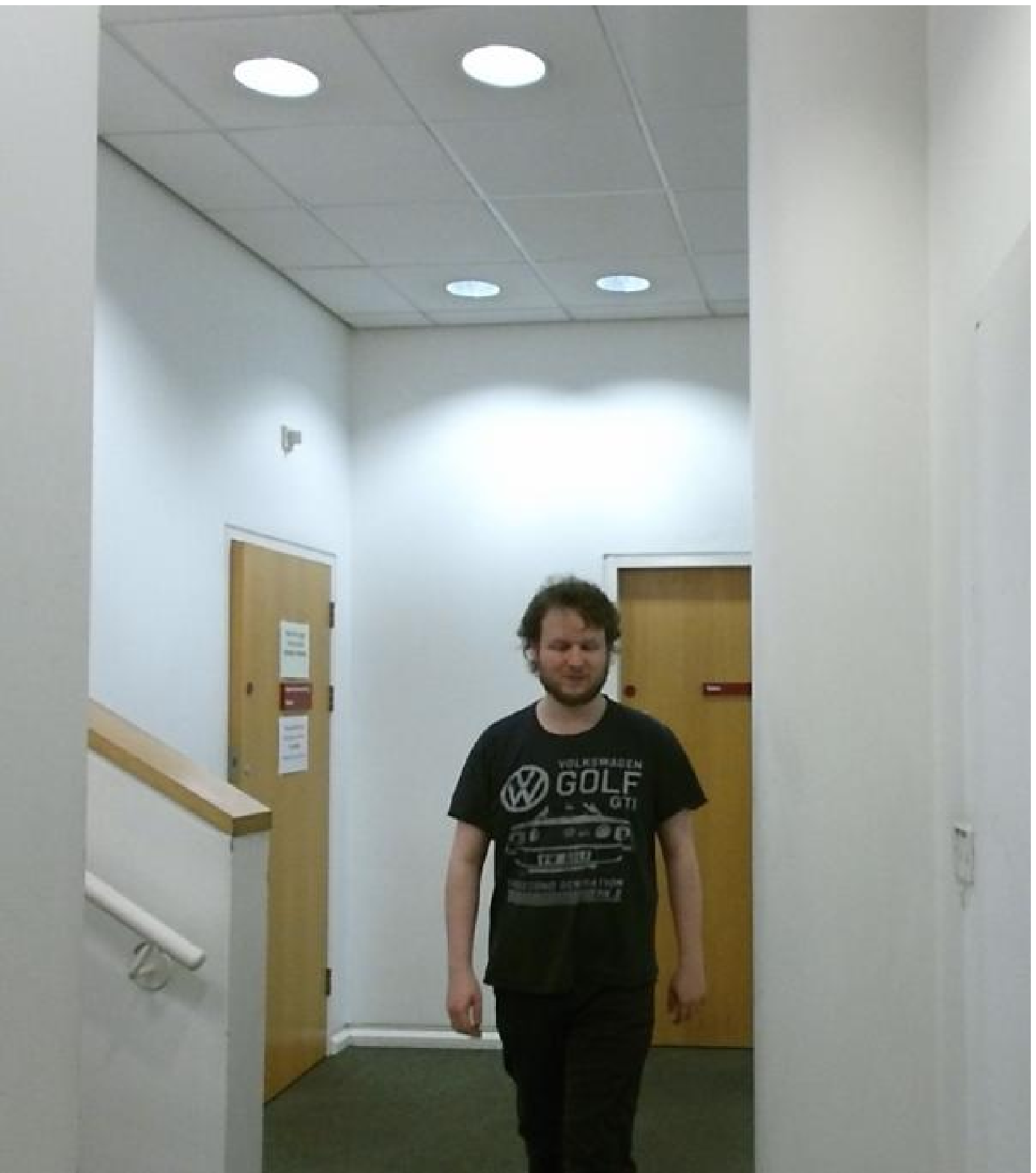}}\hspace{-2.2pt}
\cfbox{orng}{\includegraphics[width=0.11\textwidth]{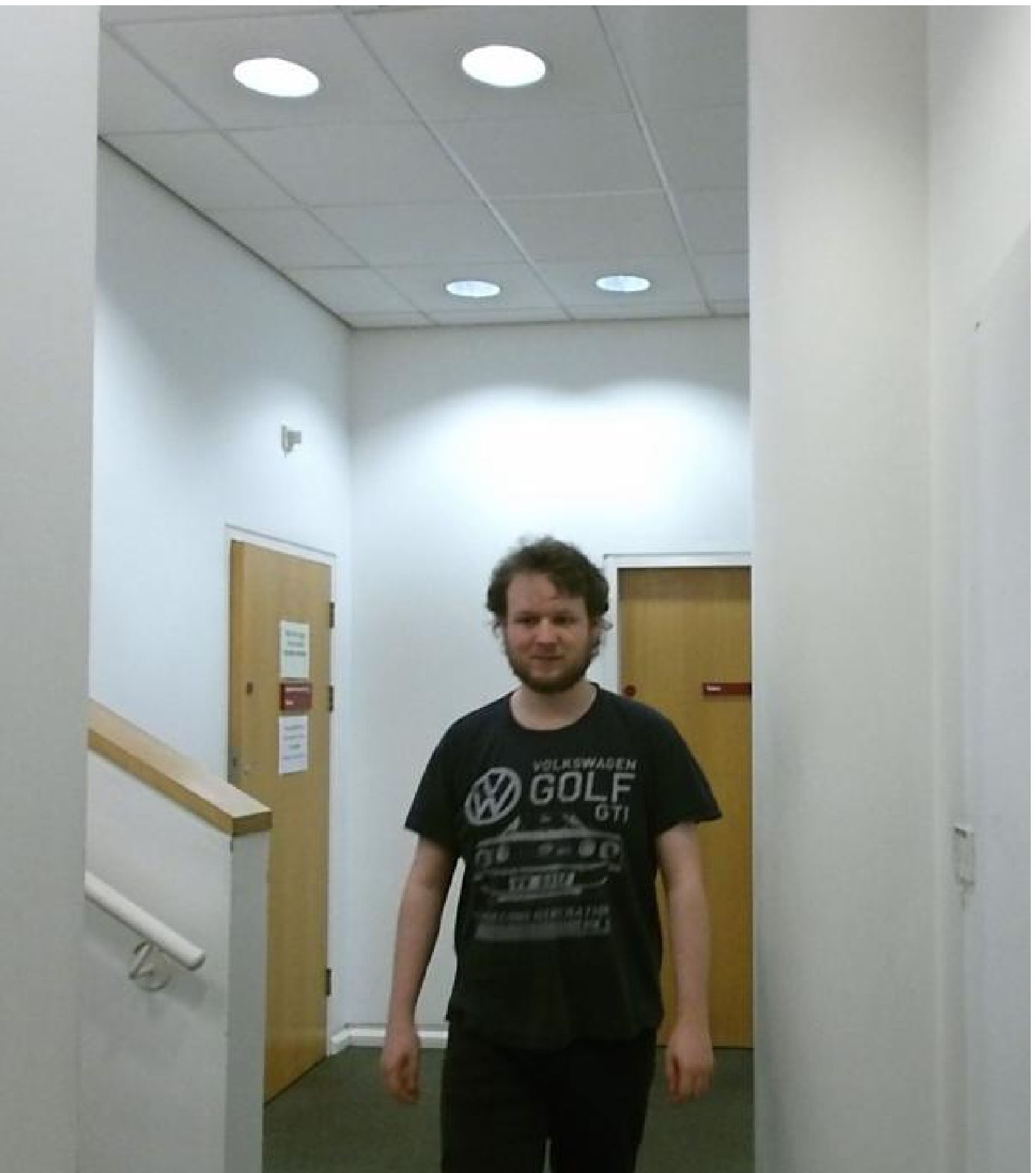}}\hspace{-2.2pt}
\cfbox{orng}{\includegraphics[width=0.11\textwidth]{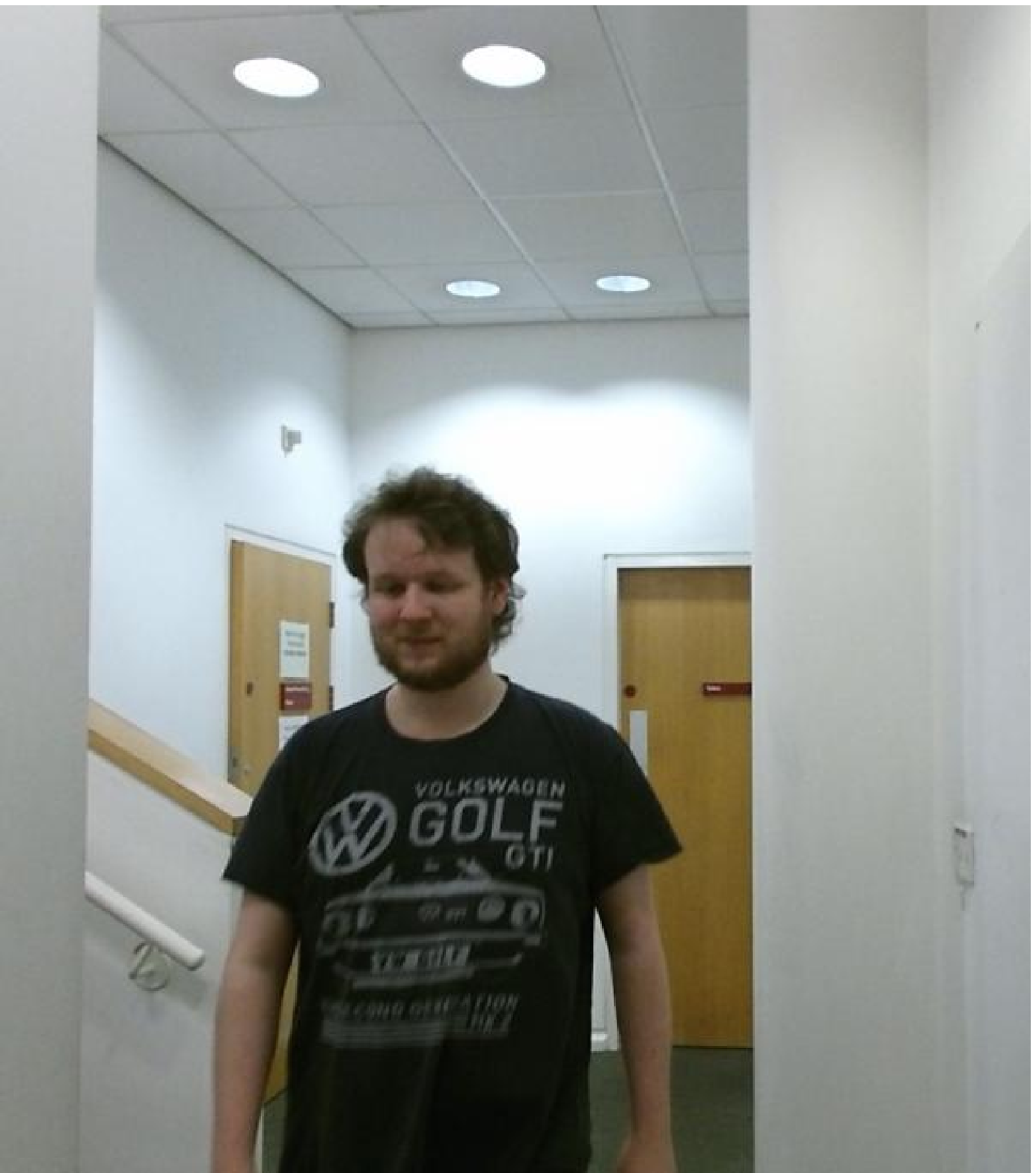}}\hspace{-2.2pt}
\cfbox{orng}{\includegraphics[width=0.11\textwidth]{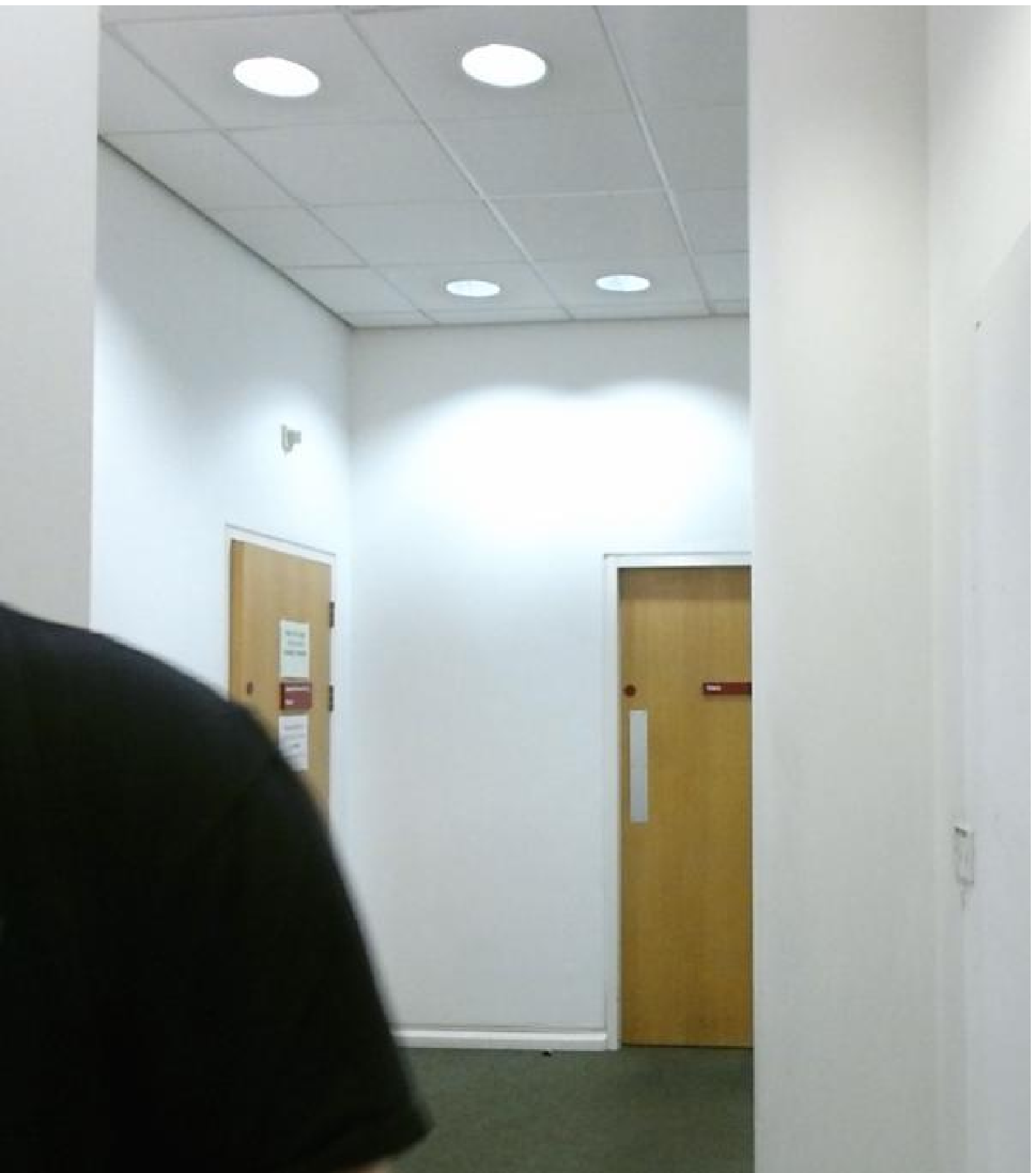}}\\

\vspace*{-6pt}
\caption{Annotation of \textit{completion moment}: Two examples per action. \textit{Pre-completion} frames are bordered in orange and \textit{post-completion} in  purple. \textbf{From Top:} HMDB \textit{pick}, UCF101 \textit{blowing candles}, RGBD-AC \textit{switch} (one complete sequence and one incomplete). }
\label{fig:switch_example}
\end{figure}

\vspace*{-4pt}
\section{Action Completion - A Moment in Time}
\label{sec:assumptions}
\vspace*{-2pt}

We first present our proposal for formulating the problem of localising an action's completion as detecting a moment in time, beyond which the action's goal is believed to be completed by a human observer. We make three reasonable assumptions: 
\vspace*{-4pt}
\begin{itemize}[leftmargin=*]
\item \textbf{Momentary Completion}: We aim to detect a single frame in the sequence - that is the first frame where a human observer would be sufficiently confident that the goal has been achieved. We refer to frames prior to the \textit{completion moment} as {\textit{pre-completion}} frames, and those {from the \textit{completion moment} onwards} as {\textit{post-completion}} frames. 
\vspace*{-3pt}
\item \textbf{Temporally Segmented Sequences:} 
We assume that the action is attempted during each sequence, in train or test, at least once, but not necessarily completed. We aim to detect the first \textit{completion moment} per sequence, if at all, or label the attempt as being incomplete. 
\vspace*{-3pt}
\item \textbf{Consistent Labeling}: For each action, we assume annotators are given a non-ambiguous definition of the \textit{completion moment}, so all train and test sequences are labeled consistently. For example, in the action \textit{`blowing candle'}, the consistent label for the \textit{completion moment} should indicate the moment when the flames of all candles go out. Note that the proposed model is independent of the definition of the \textit{completion moment} per action. It only assumes the moment is consistently labeled across sequences. 
\end{itemize}
\vspace*{-1pt}



Figure~\ref{fig:switch_example} shows sample sequences, labeled with \textit{completion moments}, for three actions from the various datasets we annotate and use:
(i) \textit{pick} from HMDB, where the \textit{completion moment} is when the object is lifted off the surface (ii) \textit{blowing candles} from UCF101, where the \textit{completion moment} is when all the candles are blown out  and (iii) \textit{switch light} from RGBD-AC, where \textit{completion moment} is when the room's {illumination} changes.

Labeled sequences for a given action are the input to our method, presented next.
For each sequence $i$, one \textit{completion moment} is labeled {if present}, which we refer to as $\tau_i$, such that $1 \le \tau_i \le T_i$, and $T_i$ is the sequence length, {or the sequence is labeled as incomplete.} 



\vspace*{-6pt}
\section{Temporal Model for Moment Detection}
\label{sec:ac_model}
\vspace*{-4pt}

{To detect the \textit{completion moment} within a sequence}, one could naively attempt to train a classifier that {singularly} separates the frame indicating the \textit{completion moment} from the rest of the video. However, evidence for the \textit{completion moment} can be collected from all (or any) {frames} in the sequence. Take for example the action `\textit{pick}'; the pose of the person is likely to change and evolve as they approach the object to be picked, and similarly observing the object in hand as the hand retracts gives further support for completion. We propose a temporal model that learns local (i.e.~frame-level) predictions, within a recurrent neural network, towards global (i.e.~sequence-level) detection, trainable end-to-end.

Our proposed temporal model is a Convolutional-Recurrent Neural Network. {We} 
describe the frame-level voting nodes in Sec~\ref{sub:vote} {and then} show how the unfolded temporal model, over a sequence, can accumulate votes towards moment detection in Sec~\ref{sec:sequence}.

\vspace*{-6pt}
\subsection{Frame-level Voting Recurrent Node}
\label{sub:vote}
\vspace*{-4pt}


Each frame in the sequence, whether prior to the \textit{completion moment}, or post completion, could contribute to the \textit{completion moment} detection.
We refer to this contribution as `voting', i.e. a frame can vote for when the action will be (or has been) completed. 
Two ways are proposed in which such voting can take place:
\vspace*{-5pt}
\begin{enumerate}[leftmargin=*]
\item Classification Voting: At each time step $t$, the sequence is split into two parts: ${[1 \cdots t]}$ and ${[t+1 \cdots T]}$, where $T$ is the sequence length. The classification vote primarily distinguishes the split within which the \textit{completion moment} resides.
\vspace*{-5pt}
\item Regression Voting: At each time step $t$, the relative position of the \textit{completion moment} is predicted, normalised to allow for sequences of various lengths.
\end{enumerate}}
\vspace*{-5pt}

Figure~\ref{fig:single-frame} shows the architecture of our proposed frame-level voting recurrent node, which can be used to predict both the classification and regression votes defined above, trained using a joint classification-regression loss function. Each input frame is passed through convolutional, pooling and fully connected layers. Then, an LSTM layer combines past information with the current observation. The LSTM output $h_t$ is trained to perform frame-level classification as well as frame-level regression as follows:

\begin{figure}[t]
\centering
\includegraphics[width=1\textwidth]{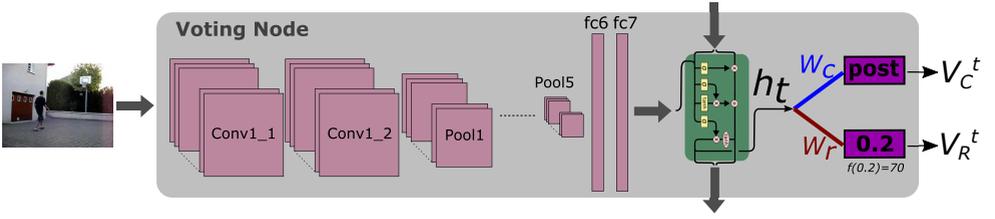}
\vspace*{-14pt}
\caption{The input image passes through convolutional, pooling and fully-connected layers, and then an LSTM cell to capture temporal dependencies from the past. The node outputs classification $V_C^t$ and regression $V_R^t$ votes for the \textit{completion moment}.}

\label{fig:single-frame}
\end{figure}


\noindent \textbf{Frame-Level Classification Voting ($V_C^t$):}
To decide whether the \textit{completion moment} is before or after the current time step $t$, we primarily need to predict whether the current observation is \textit{pre-} or \textit{post-} completion.
We thus train for $V_C^t$ by classifying the current observation, using a Sigmoid cross-entropy loss function on top of the LSTM hidden layer, such that
\begin{eqnarray}
&C_t &= g(W_c h_t+B_c) \label{eq:Ct}  ~ , \\
&L_{C_t} &=-(y_{t}log(C_t)+(1-y_{t})log(1-C_t)) \label{eq:Lc}  ~ , 
\end{eqnarray}
\noindent where  $W_c$ and $B_c$ are the weights and biases for classification, respectively, {$g$ is the Sigmoid activation function} and $y_t$ is the supervised label.
The \textit{pre-} and \textit{post-} class labels are assigned to all frames $t < \tau$ and $t \geq \tau$, respectively; sequence subscript is removed for simplicity. 

The classifier then allows to vote for the presence of the \textit{completion moment} in one of two splits of the sequence, namely $[1 \cdots t]$ or {$[t+1 \cdots T]$}. 
Specifically, if the observation at time $t$ is classified as being \textit{pre-}completion, then the \textit{completion moment} is believed to be within {$[t+1 \cdots T]$, or could be incomplete. 
{To account for incompletion, we extend the end of the second split to $T+1$, to allow votes to be cast for an incomplete sequence, so the second split becomes {$[t+1 \cdots T+1]$}.}
Otherwise, the \textit{completion moment} is believed to be within $[1 \cdots t]$.
The classification vote contributes equally to voting within the split.
We define $V_C^t$ as a one-dimensional vector of length $T+1$, representing the vote assigned to all frames in the sequence.
For each frame $j$, the vote cast by the current frame $t$, $V_C^t(j)$, is 
\begin{equation}\label{eq:V_c}
V_{C}^t(j)=\begin{cases}
    \frac{1}{T-t+1} & j > t \ \ \land \ \ C_t=\text{pre}\\
    \frac{1}{t} & j \le t \ \ \land \ \ C_t=\text{post}\\
    0 & \text{otherwise}
  \end{cases} \qquad \qquad \forall j = 1 \cdots T+1 . 
\end{equation}
The frame-level classification votes $V_C^t$ are then accumulated 
(see Sec.~\ref{sec:sequence}).

\noindent \textbf{Frame-Level Regression Voting $(V_R^t)$:}
While $V_C^t$ assigns an equal vote to all frames {within each} of the splits in the sequence, defined by $t$, regression voting $V_R^t$ provides stronger evidence that can localise the \textit{completion moment}, by predicting its relative position to $t$. 
This relative position encapsulates the \textit{remaining} time to or \textit{elapsed} time from the \textit{completion moment}. 
We compute the relative time as that between $t$ and the \textit{completion moment} $\tau$, normalised by $\tau$, i.e. $\frac{t-\tau}{\tau}$. This provides a more robust relative temporal position than the alternative $\frac{t-\tau}{T}$ {which would} 
differ with the length of the sequence. Note that this value is negative during pre-completion, that is $t < \tau$.

To train for frame-level regression, the hidden output $h_t$ in the voting recurrent node learns to predict the relative time, using a Euclidean loss function, to obtain

\vspace*{-16pt}
\begin{eqnarray}\label{eq:Lr}
&R_t &=W_r h_t + B_r ~  , \\
&L_{R_t} &=(R_t-\frac{(t - \tau)}{\tau})^2  ~ , 
\end{eqnarray}  
\vspace*{-16pt}

\noindent where $W_r$ and $B_r$ are the weights and biases for regression, respectively.
$R_t$ can then be used to predict the \textit{completion moment} {at the corresponding time $t$} as
$f(t, R_t) = \frac{t}{R_t+1}$.


Similar to classification voting, we define $V_R^t$ as a one-dimensional voting vector, and 
use a Gaussian with uncertainty $\sigma$ around the predicted \textit{completion moment}, $f(t, R_t)$, such that
\begin{equation}\label{eq:V_r}
 \qquad V_{R}^t(j) =\beta e^{\frac{-(j-f(t,R_t))^2}{2\sigma^2}}  \qquad \qquad \forall j = 1 \cdots T+1,
\end{equation} 
where $\beta$ represents the inverse of the selected {area under curve} of the Gaussian. Experimentally, we only {compute} the regression vote within a window of size $\alpha T$,
in order to reduce the complexity of calculating the vote for all time steps in the sequence.

\noindent \textbf{Training Loss:} As a forward recurrent neural network, we can then train all parameters using a combined loss on all sequences and their frames, specified as,

\vspace*{-12pt}
\begin{equation}
L = \sum_i \bigl(\frac{1}{T_i} \sum_{t_i=1}^{T_i} (L_{C_{t_i}} + L_{R_{t_i}})\bigr) ,
\label{eq:total_loss}
\end{equation}
\vspace*{-12pt}

\noindent allowing all sequences to contribute equally to the loss function regardless of the sequence length.
The loss is propagated back through the recurrent voting nodes.

\vspace*{-4pt}
\subsection{Sequence-level prediction of \textit{completion moment}}
\label{sec:sequence}
\vspace*{-4pt}

The votes by individual frames are accumulated to make sequence-level predictions of the \textit{completion moment}. 
Note that we do not propagate ambiguity in the decisions of the individual frames, and assume each frame is equally certain about its votes.
Other approaches that {could} integrate frame voting uncertainty, or learn temporal attention, are left for future investigation.
We focus on assessing the robustness of using the classification vs the regression votes as follows:
\textbf{(i) Classification\textsubscript{\textit{pre}}-Classification\textsubscript{\textit{post}} (C-C)}: all frames use classification-based voting, 
\textbf{(ii) Regression\textsubscript{\textit{pre}}-Regression\textsubscript{\textit{post}} (R-R):} all frames vote using their regression-based voting, \textbf{(iii) Regression\textsubscript{\textit{pre}}-Classification\textsubscript{\textit{post}} (R-C):} frames classified as \textit{pre-completion} use their regression-based voting, while \textit{post-} frames use classification-based voting, and correspondingly \textbf{(iv) Classification\textsubscript{\textit{pre}}-Regression\textsubscript{\textit{post}} (C-R)}. {Symbolically,}


\begin{minipage}{0.3\textwidth}
\begin{eqnarray}
&V_{C-C} &= \sum_t V_C^t\\
&V_{R-R} &= \sum_t V_R^t
\end{eqnarray}
\end{minipage}
\begin{minipage}{0.6\textwidth}
\begin{eqnarray}
&V_{R-C} &= \sum_{t: C_t = \text{pre}} V_R^t + \sum_{t: C_t = \text{post}} V_C^t\\
&V_{C-R} &= \sum_{t: C_t = \text{pre}} V_C^t + \sum_{t: C_t = \text{post}} V_R^t
\end{eqnarray} \end{minipage}

\vspace*{12pt}

\noindent Fig.~\ref{fig:model} illustrates the various approaches to frame-based votes.
The predicted sequence-level \textit{completion moment} $\tau^p$ is then the frame with the maximum accumulative vote: 
\vspace*{-6pt}
\begin{equation}\label{eq:cp}
\tau^p= \arg\max_j V(j) .
\end{equation}
\begin{figure}[t]
\centering
\includegraphics[width=1\textwidth]{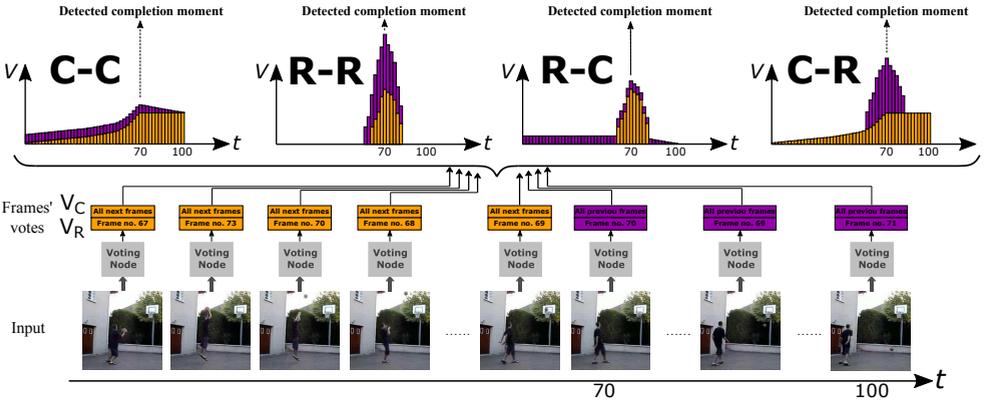}
\vspace*{-6pt}
\caption{Sequence-level completion detection by accumulating frames' votes. {The schemes use classification and/or regression voting. Sample sequence from action \textit{basketball}.}}
\label{fig:model}
\end{figure}

\vspace*{-20pt}
\section{Experiments and Results}
\label{sec:res}
\vspace*{-2pt}


\noindent \textbf{Dataset and Completion Annotation:}
To show the generality of our work, we select 16 actions from 3 public datasets, and annotate them for their \textit{completion moments}. 
We avoid actions for which completion would be difficult to define or it just marks the end of the action, \textit{e.g. run, play piano, laugh}. However, we select actions that cover both sports-based and daily actions. 
For each sequence, we provide an annotation of the first \textit{completion moment}, by a single annotator\footnote{Annotations available at: \url{https://github.com/FarnooshHeidari/CompletionDetection}.}.

\noindent \textbf{HMDB~\cite{HMDB}:} We annotate all sequences of 5 {actions:} \textit{catch, drink, pick, pour} and \textit{throw}. In total, these are 494 sequences, of which 93.5\% are complete, i.e. the action's goal is successfully achieved. While HMDB does not aim for completion detection, a few sequences include attempts that are unsuccessful.

\noindent \textbf{UCF101~\cite{UCF101}:} We annotate all sequences of 5 {actions:} \textit{basketball, blowing candles, frisbee catch, pole vault} and \textit{soccer penalty}. These are 650 sequences, {of which} 80.5\% are complete.

\noindent \textbf{RGBD-AC~\cite{Heidari}:} We use the RGB input our previously introduced dataset~\cite{Heidari}, and annotate all 414 sequences which include 6 actions: \textit{switch, plug, open, pull, pick} and \textit{drink}, of which 50.5\% are complete. In this dataset, subjects are disrupted from completing the action, e.g. {a drawer they attempt to open is locked}.

\noindent We apply `leave-one-person-out' to evaluate the RGBD-AC dataset, while for HMDB and UCF101, the provided train and test splits are used.



\noindent \textbf{Implementation Details:} For the convolution and pooling layers, we use the spatial stream CNN from~\cite{Simonyan2014} which uses the VGG-16 architecture~\cite{Simonyan2014_CoRR}, pre-trained on UCF101.
This CNN is then fine-tuned per action, using the two classes of \textit{pre-} and \textit{post-completion} frames. For fine-tuning, 20 epochs are performed, and the learning rate is started {at $10^{-3}$, divided by 10 {at epochs 3 and 5}. 
All the other hyper-parameters are set as in~\cite{Feichtenhofer2016}. 

The 4096-dimension vector of $fc7$ forms the input to the single LSTM layer with 128 hidden units. 
Initialisation is random, trained for 10 epochs. The learning rate is $10^{-2}$ for the first 5 epochs and is fixed at $10^{-3}$ for the remaining epochs. We use a mini-batch size of one sequence {and} parameters $\alpha$, $\beta$ and $\sigma$ are set 0.1, 0.5 and 30, respectively.
While the proposed method represents an end-to-end trainable model, in the presented results, we train a CNN and feed the $fc7$ features into the LSTM. Efficient end-to-end training of the proposed temporal model is challenging using available hardware, and is left for future work.


\begin{table}[t]
\centering
\fontsize{6}{6} \selectfont{
\def\arraystretch{1.3}
\begin{tabular}{|c|c|c|c|c|c|c|c||c|c|c|c|c|c|}
\cline{3-14}
\multicolumn{2}{c|}{ } & \multicolumn{6}{c||}{\bf Accuracy} & \multicolumn{6}{c|}{$RD$}\\ \cline{3-14}
\multicolumn{2}{c|}{ } & \textbf{Pre-V} & \textbf{$V_R^T$} & \textbf{C-C} & \textbf{R-R} & \textbf{R-C} & \textbf{C-R} & \textbf{Pre-V} & \textbf{$V_R^T$}  & \textbf{C-C} & \textbf{R-R} & \textbf{R-C} & \textbf{C-R}\\ \hline
\multirow{6}{*}{\rotatebox[origin=c]{90}{HMDB}} & \textit{catch} & 77.3 & 79.1 & 75.9 & 80.5 & 76.7 & \textbf{82.3} & 0.23 & 0.21 & 0.24 & 0.20 & 0.23 &\textbf{0.18}\\ \cline{2-14}
&\textit{drink} & 77.3 & 69.3 & 73.2 & 78.0 & 75.9 & \textbf{80.5} & 0.21 & 0.31 & 0.27 & 0.22 & 0.24 & \textbf{0.19} \\ \cline{2-14}
&\textit{pick} & 80.6 & 79.5  & 79.7 & 79.9 & 74.7 & \textbf{84.2} & 0.22 & 0.20 & 0.20 & 0.20 & 0.25 & \textbf{0.16}\\ \cline{2-14}
&\textit{pour} & 76.5 & 68.3  & 71.1 & 80.0 & 78.7 & \textbf{81.2} & 0.23 & 0.32 & 0.29 & 0.20 & 0.21 & \textbf{0.19}\\ \cline{2-14}
&\textit{throw} & 68.7 & 74.3 & 63.4 & 74.6 & 65.8 & \textbf{80.4} & 0.32 & 0.26 & 0.37 & 0.25 & 0.34 & \textbf{0.20}\\ 
\hline \hline

\multirow{6}{*}{\rotatebox[origin=c]{90}{UCF101}} & \textit{basketball} & \textbf{86.5} & 78.0 & 84.5 & 79.5 & 79.1 & 85.1 & 0.21 & 0.22 & 0.16 & 0.20 & 0.21 & \textbf{0.15} \\ \cline{2-14}
& \textit{blowing candles} & 86.8 & 88.3 & 86.4 & 84.2 & 78.2 & \textbf{90.9} & 0.16 & 0.12 & 0.14 & 0.16 & 0.22 & \textbf{0.09} \\ \cline{2-14}
& \textit{frisbee catch} & 81.7 & 84.1  & 80.3 & 78.3 & 74.6 & \textbf{85.9} & 0.24 & 0.16 & 0.20 & 0.22 & 0.25 & \textbf{0.14} \\ \cline{2-14}
& \textit{pole vault} & 85.0 & 83.3 & 82.6 & 88.4 & 80.1 & \textbf{90.6} & 0.19 & 0.17 & 0.17 & 0.12 & 0.20 & \textbf{0.09} \\ \cline{2-14}
& \textit{soccer penalty} & 85.5 & 86.6 & 85.8 & 87.1 & 85.6 & \textbf{88.5} & 0.15 & 0.13 & 0.14 & 0.13 & 0.14 & \textbf{0.11} \\ 
\hline \hline

\multirow{6}{*}{\rotatebox[origin=c]{90}{RGBD-AC}} & \textit{switch} & \textbf{99.9} & 93.9 & \textbf{99.9} & 98.1 & 92.7 & 98.9 & \textbf{0.00} & 0.06 & \textbf{0.00} & 0.02 & 0.07 & 0.01 \\ \cline{2-14}
& \textit{plug} & 98.3 & 93.2 & \textbf{98.5} & 96.1 & 93.0 & 97.2 & 0.02 & 0.07 & \textbf{0.01} & 0.04 & 0.07 & 0.03 \\ \cline{2-14}
& \textit{open} & \textbf{91.1} & 86.1 & \textbf{91.1} & 86.7 & 80.4 & 89.9 & 0.12 & 0.14 & \textbf{0.09} & 0.13 & 0.20 & 0.10 \\ \cline{2-14}
& \textit{pull} & 97.7 & 89.1 & \textbf{97.8} & 94.1 & 91.5 & 97.0 & 0.10 & 0.11 & \textbf{0.02} & 0.06 & 0.08 & 0.03 \\ \cline{2-14}
& \textit{pick} & 91.5 & 89.1 & 89.9 & 93.2 & 83.6 & \textbf{95.0} & 0.11 & 0.11 & 0.10 & 0.07 & 0.16 & \textbf{0.05} \\ \cline{2-14}
& \textit{drink} & 88.6 & 79.0 & 85.3 & 90.9 & 85.8 & \textbf{92.1} & 0.11 & 0.21 & 0.15 & 0.09 & 0.14 & \textbf{0.08} \\ 
\hline \hline

\multicolumn{2}{|c|}{\textbf{complete}} & 82.3 & 78.1 & 79.6 & 83.1 & 77.7 & \textbf{85.6} & 0.19 & 0.22 & 0.20 & 0.17 & 0.22 & \textbf{0.14}\\ \hline

\multicolumn{2}{|c|}{\textbf{incomplete}} & 93.4 & 94.8 & 94.3 & 90.4 & 88.8 & \textbf{96.1} & 0.13 & 0.05 & 0.06 & 0.10 & 0.11 & \textbf{0.04}\\ \hline

\multicolumn{2}{|c|}{\textbf{total}} & 85.0 & 82.2 & 83.2 & 84.9 & 80.4 & \textbf{88.1} & 0.17 & 0.18 & 0.17 & 0.15 & 0.20 & \textbf{0.12}\\ \hline
\end{tabular}}
\caption{Results on all 16 actions, comparing frame-level classification, last-frame regression and the four sequence-level voting schemes.}
\label{tab:voting_methods}
\end{table}
\noindent \textbf{Evaluation Metrics:}
We assess the proposed model using two evaluation metrics, (i)~Accuracy: for every sequence, we compute the ratio of frames that are consistently labeled as \textit{pre-} or \textit{post-} the \textit{completion moment}, given the predicted $\tau_i^p$ and labeled $\tau_i^g$ moments,

\vspace*{-8pt}
\begin{equation}
\text{Accuracy} = \frac{1}{M} \sum_{i=1}^M \ \frac{1}{T_i} \sum_{t_i} \ (t_i < \tau_i^p \ \land \ t_i < \tau_i^g) \lor (t_i \ge \tau_i^p \ \land \ t_i \ge \tau_i^g) ~ ,
\end{equation}
\vspace*{-8pt}

\noindent where $M$ is the number of sequences, 
and (ii) the average {relative distance }error {(RD)} in predicting the \textit{completion moment}, 

\vspace*{-8pt}
\begin{equation}\label{eq:rel_dist}
RD = \frac{1}{M}\sum\limits_{i = 1}^M{\frac{||\tau_i^p-\tau_i^g||}{T_i}} \ .
\end{equation}
\vspace*{-8pt}


\begin{figure}[t]
\centering
\includegraphics[width=0.95\textwidth]{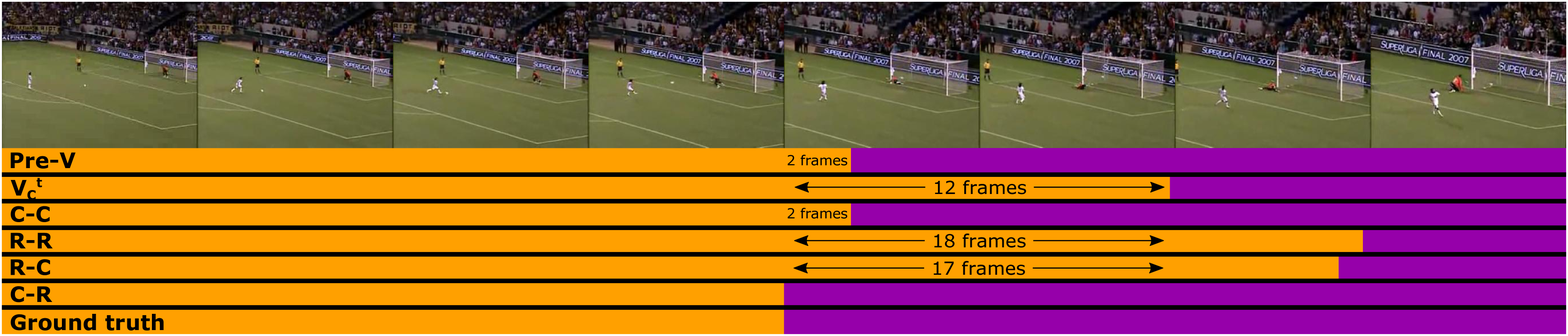}\hspace{-2.2pt}\\
\includegraphics[width=0.95\textwidth]{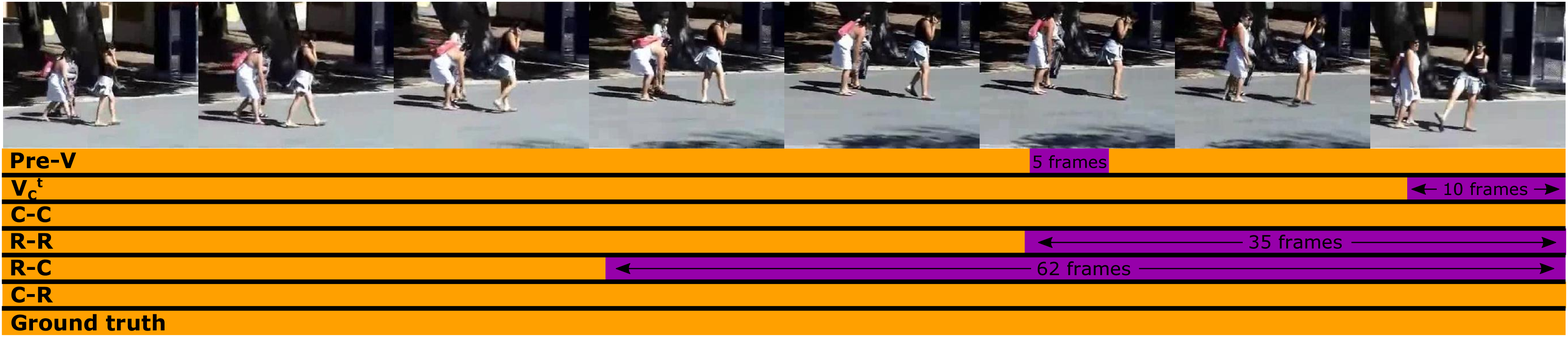}\hspace{-2.2pt}\\
\includegraphics[width=0.95\textwidth]{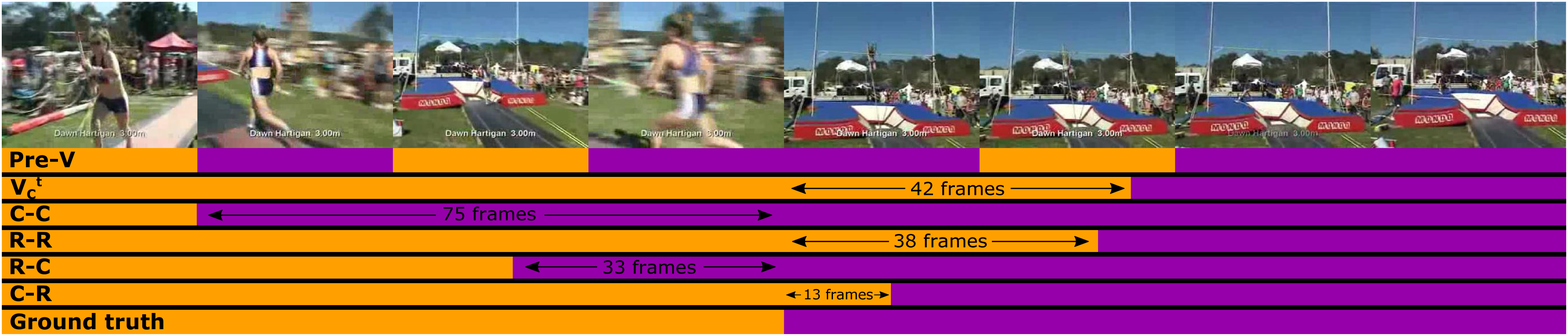}\hspace{-2.2pt}\\
\includegraphics[width=0.95\textwidth]{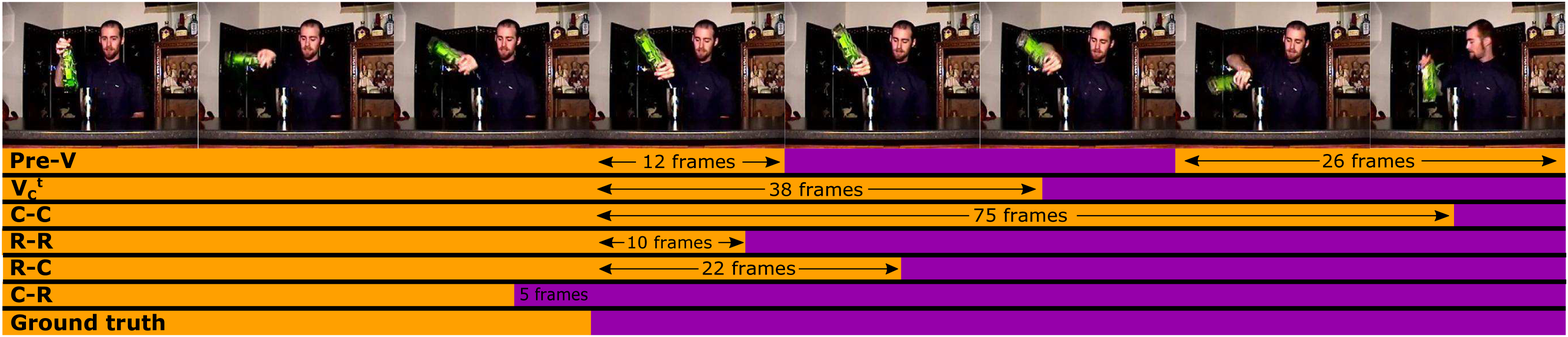}\hspace{-2.2pt}\\
\caption{Sample results for four sequences: \textit{soccer penalty, pick, pole vault} and \textit{pour}.}
\label{fig:res}
\end{figure}

\noindent \textbf{Results:}
In Table~\ref{tab:voting_methods}, 
C-R voting outperforms R-R and R-C in all actions of the three datasets. It also outperforms C-C for most actions. 
{This outcome shows that \textit{pre-completion} frames, while confident about the \textit{completion moment} being later, are unable to robustly predict the remaining time to completion.}
In contrast, a \textit{post-completion} frame which has indeed observed the completion moment is able to have a more reliable prediction of its relative position via regression. 
This also explains the poor results of method R-C in which only \textit{pre-completion} frames use regression-based voting.

In Table \ref{tab:voting_methods}, we also show two baselines: 
\textbf{(i) Pre-Voting} (Pre-V): the classification output of the LSTM hidden layer is used solely without voting. This frame-level result can have fluctuations as shown in Fig.~\ref{fig:res}. In this case, we use the first predicted \textit{post-completion} frame as $\tau_i^p$.
For HMDB and UCF101 datasets, the proposed method outperforms the frame-level classification, while for RGBD-AC, they perform comparably.
{This is because the RGBD-AC dataset is captured in one environment with a single viewpoint and thus the frame-level classifications tend to generalise easily to new sequences.}
Note that for action \textit{basketball} from UCF101, while Pre-V performs highly on the accuracy evaluation metric, the $RD$ error is higher than that of our proposed method. 
\textbf{(ii) Last-frame regression ($V_R^T$):} We only use the regression vote of the last frame. As a forward RNN is used one might question whether the accumulated result at the end of the sequence is sufficient. We show that this result is less robust than accumulating votes from all frames.
Table~\ref{tab:voting_methods} also summarises the results of complete and incomplete test sequences separately. Further action-specific results for complete and incomplete sequences are included in Appendix.

Four {qualitative} examples are presented in Fig.~\ref{fig:res}. {(1) For \textit{soccer penalty}, only C-R matches exactly the ground-truth with Pre-V and C-C giving comparable results. Using regression-voting for pre-completion frames negatively affects the \textit{completion moment} detection. (2) An incomplete \textit{pick} is correctly recognized by both C-R and C-C voting methods. (3) For \textit{pole vault}, fluctuating frame-level classifications are shown. C-R provides the closest estimation for the \textit{completion moment}. (4) For \textit{pour}, the \textit{completion moment} when the liquid is poured is predicted 5 frames earlier when using C-R voting, compared to 10 frames when R-R is used \footnote{Video of results is available at: \url{https://youtu.be/Hrxehk3Sutc}.}. 


We also present the accumulative percentage of sequences which detect the \textit{completion moment} within a certain threshold in Fig.~\ref{fig:res_plot}. We define that threshold as the absolute difference, in frames, between the predicted and ground-truth \textit{completion moments}. Results are shown for the C-R method. We correctly detect the completion moment within 1 second (30 frames) in 89\% of all test sequences, and within 0.5 second (15 frames) in {74\%} of sequences. Also \textit{completion moment} is detected for 30.4\% of sequences at the very same frame as ground-truth (i.e. 0 frame difference). Graphs are plotted for each of the 16 actions as well as the total of all actions.

\begin{figure}[t]
\includegraphics[width=0.16\textwidth]{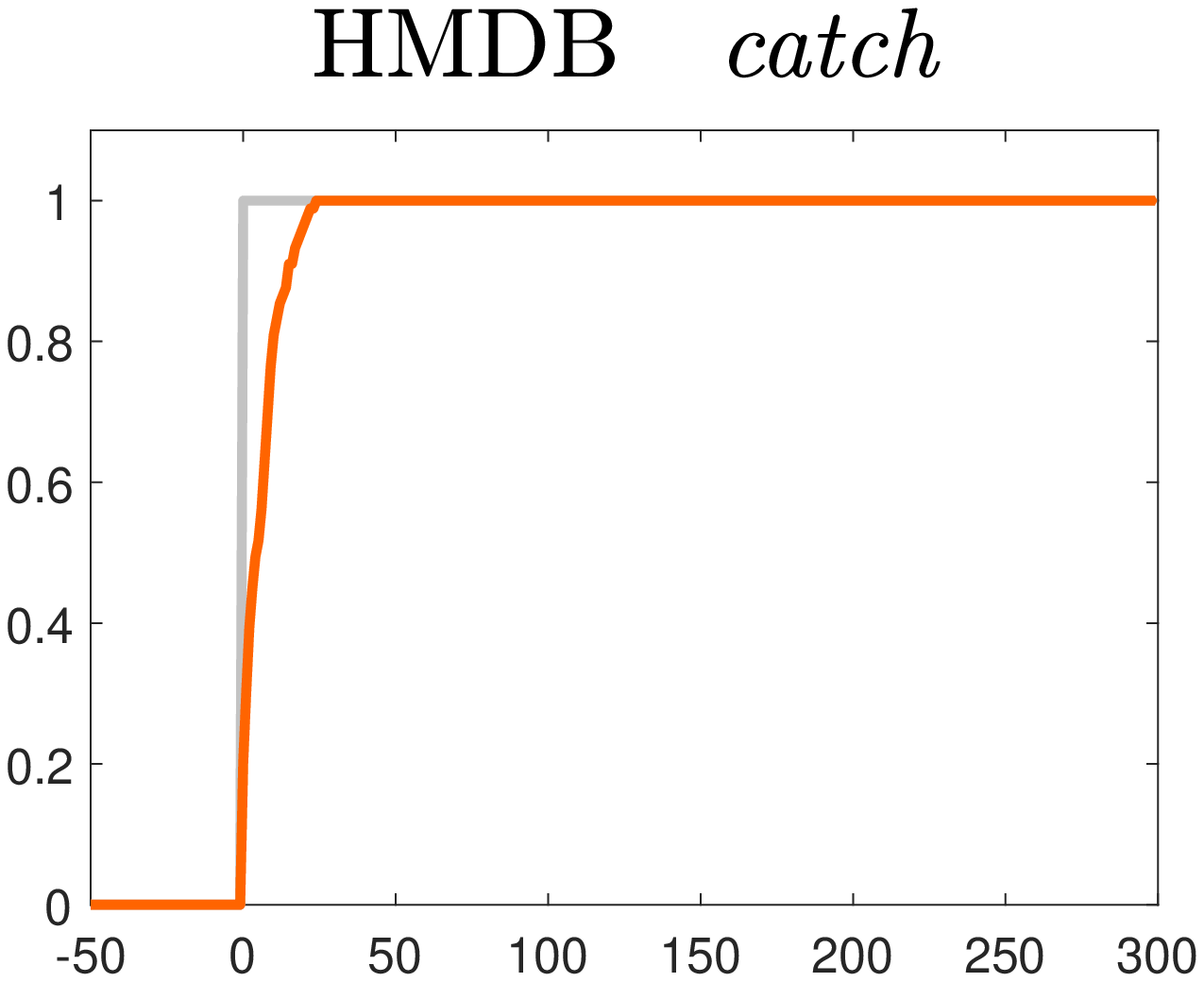}\hspace{-2.2pt}
\includegraphics[width=0.16\textwidth]{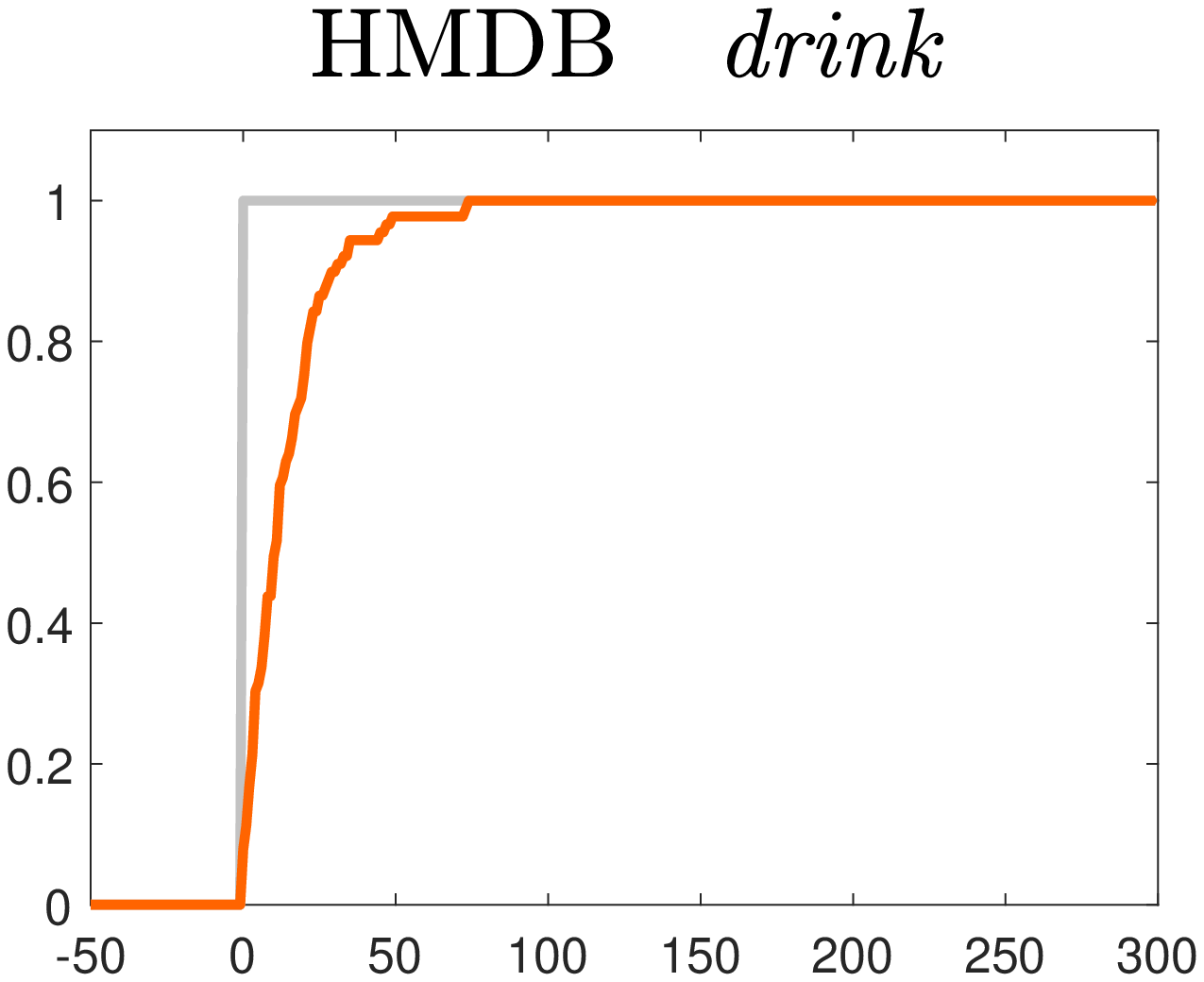}\hspace{-2.2pt}
\includegraphics[width=0.16\textwidth]{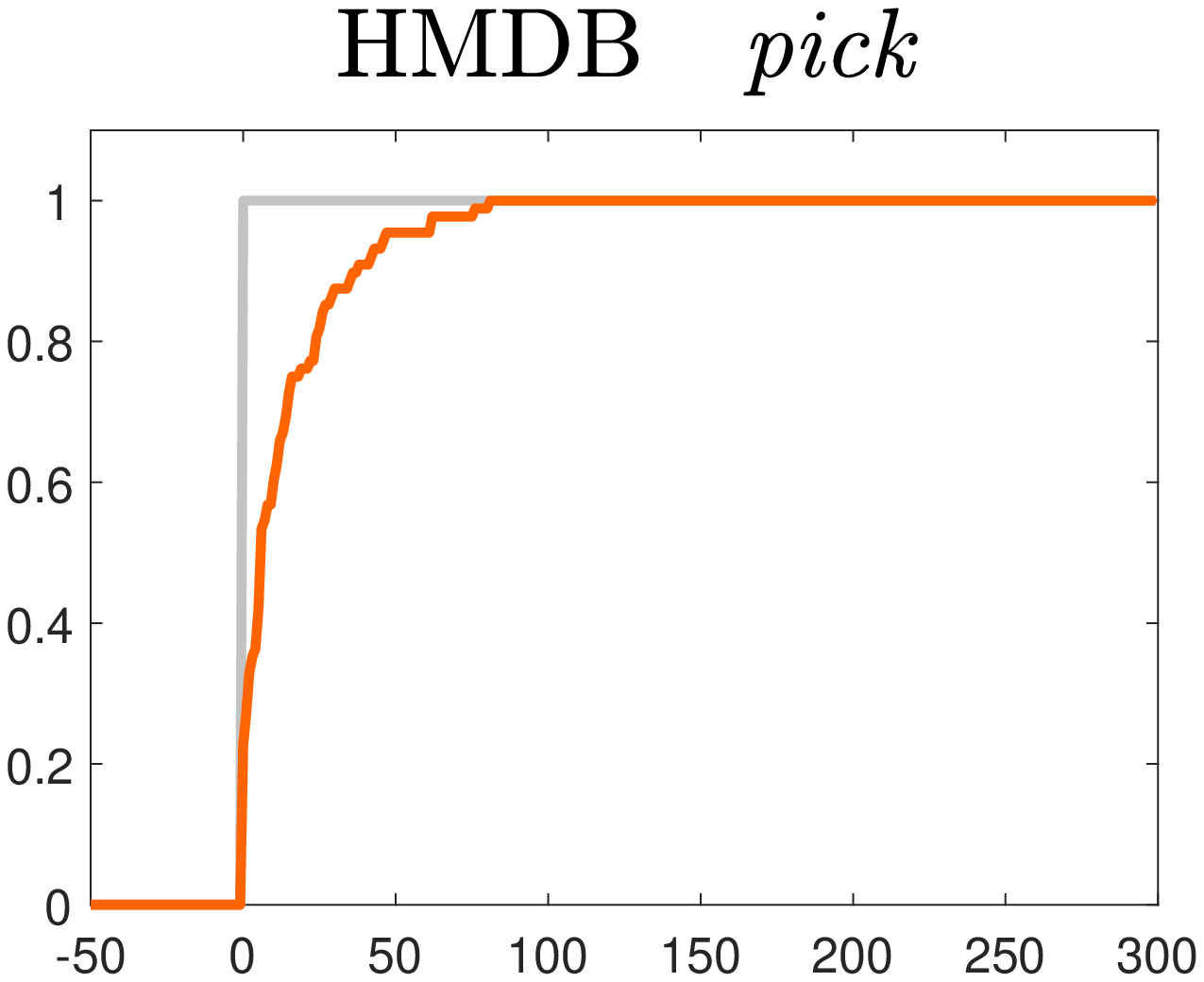}\hspace{-2.2pt}
\includegraphics[width=0.16\textwidth]{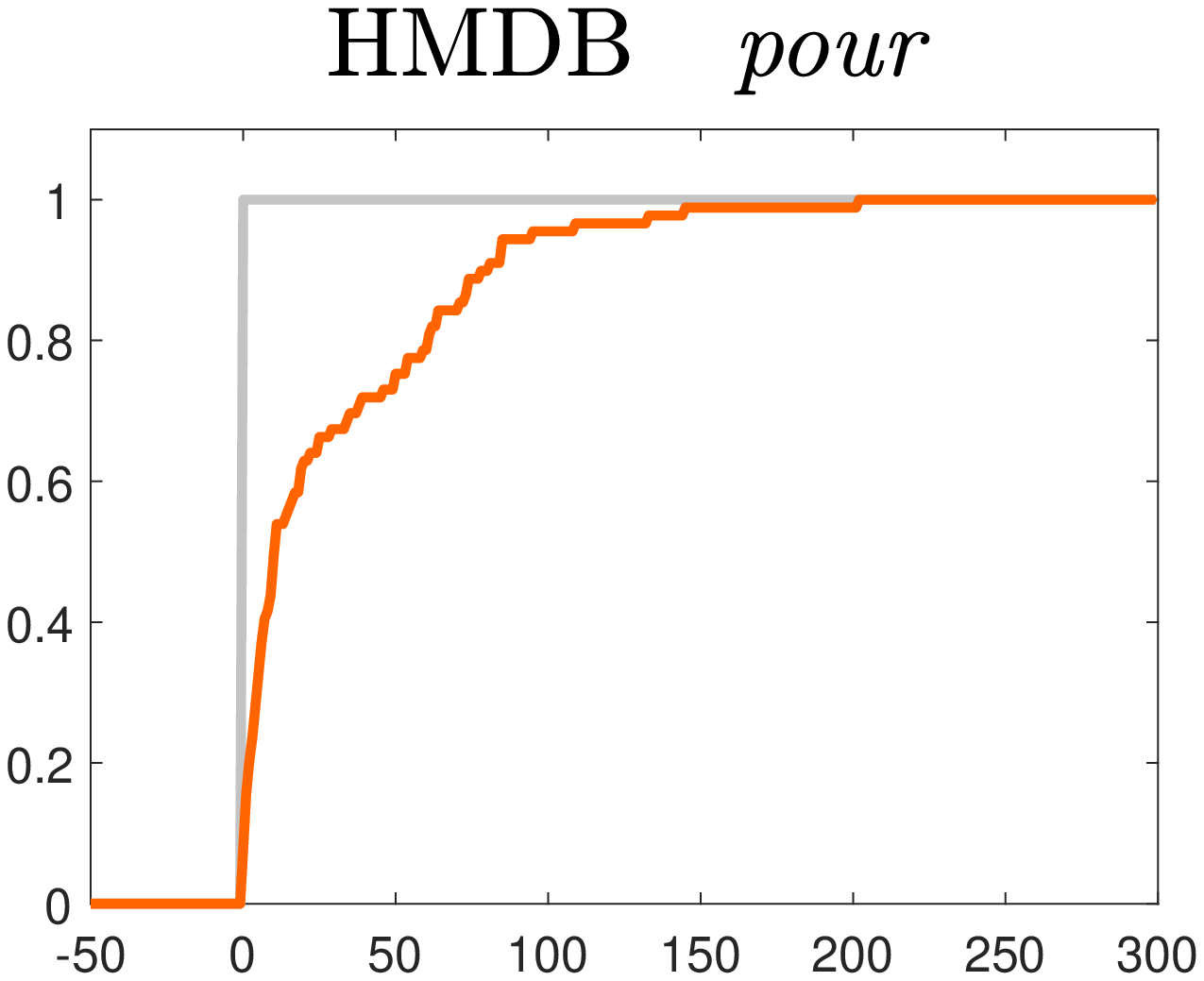}\hspace{-2.2pt}
\includegraphics[width=0.16\textwidth]{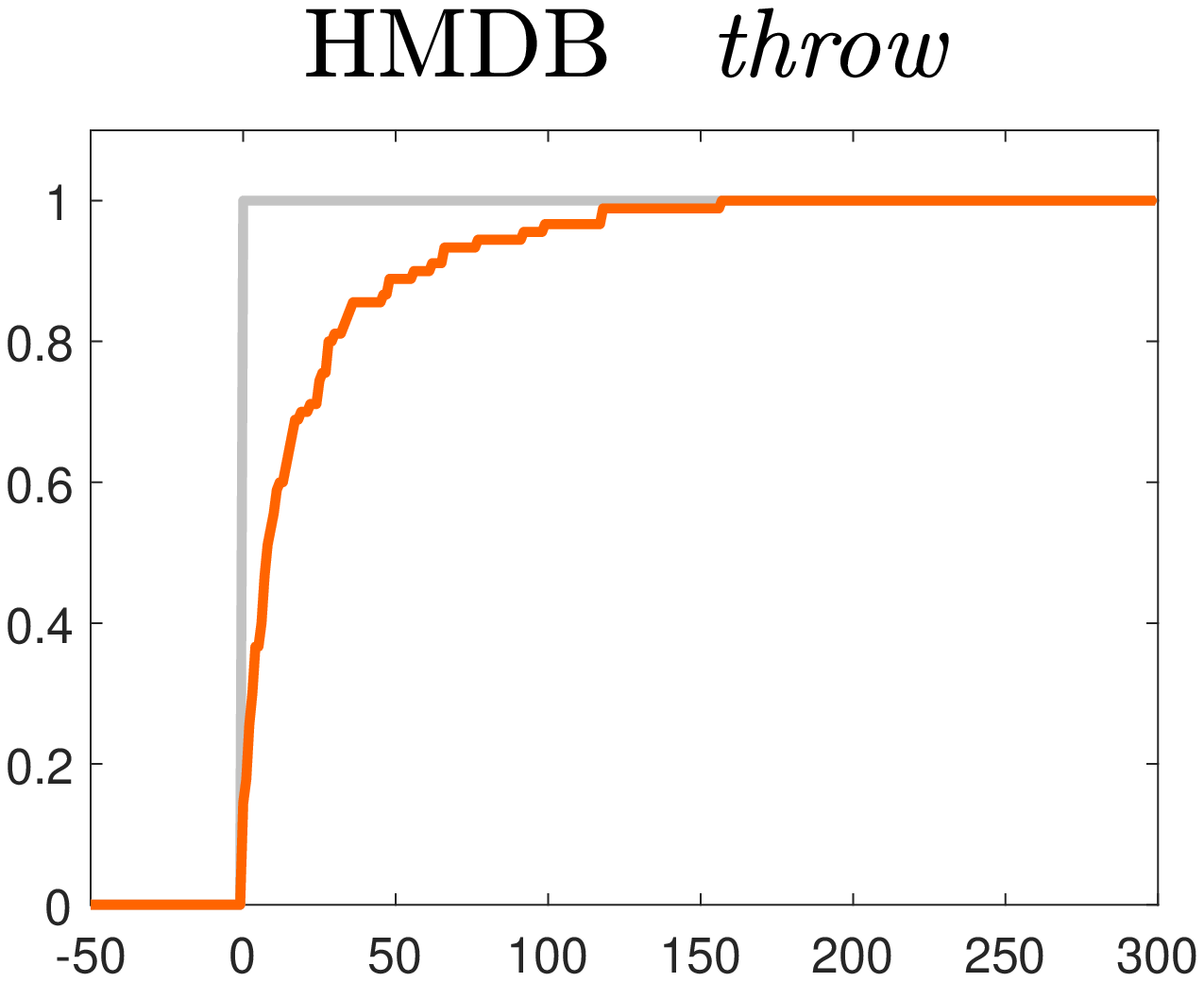}\hspace{-2.2pt}\\
\includegraphics[width=0.16\textwidth]{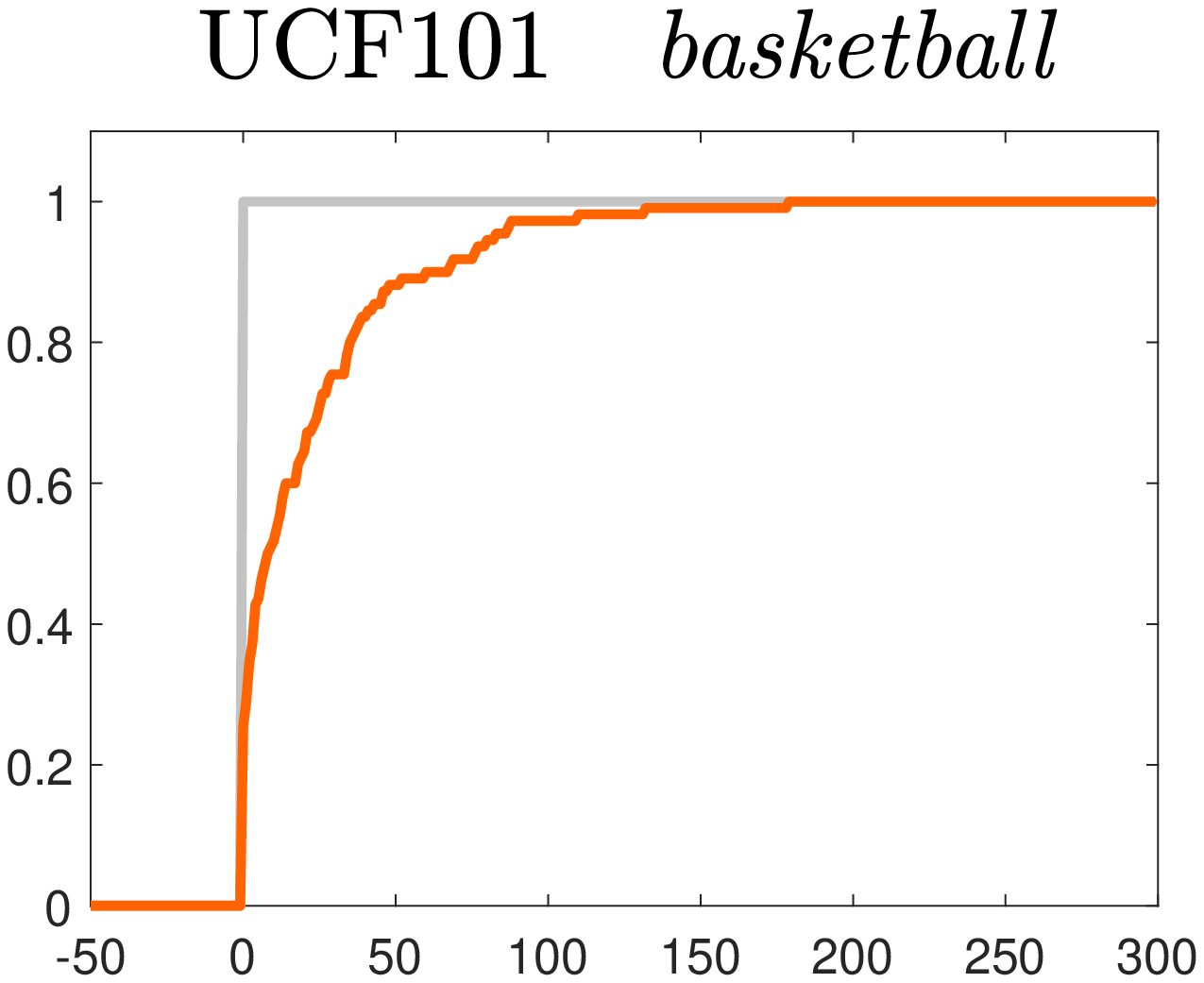}\hspace{-2.2pt}
\includegraphics[width=0.16\textwidth]{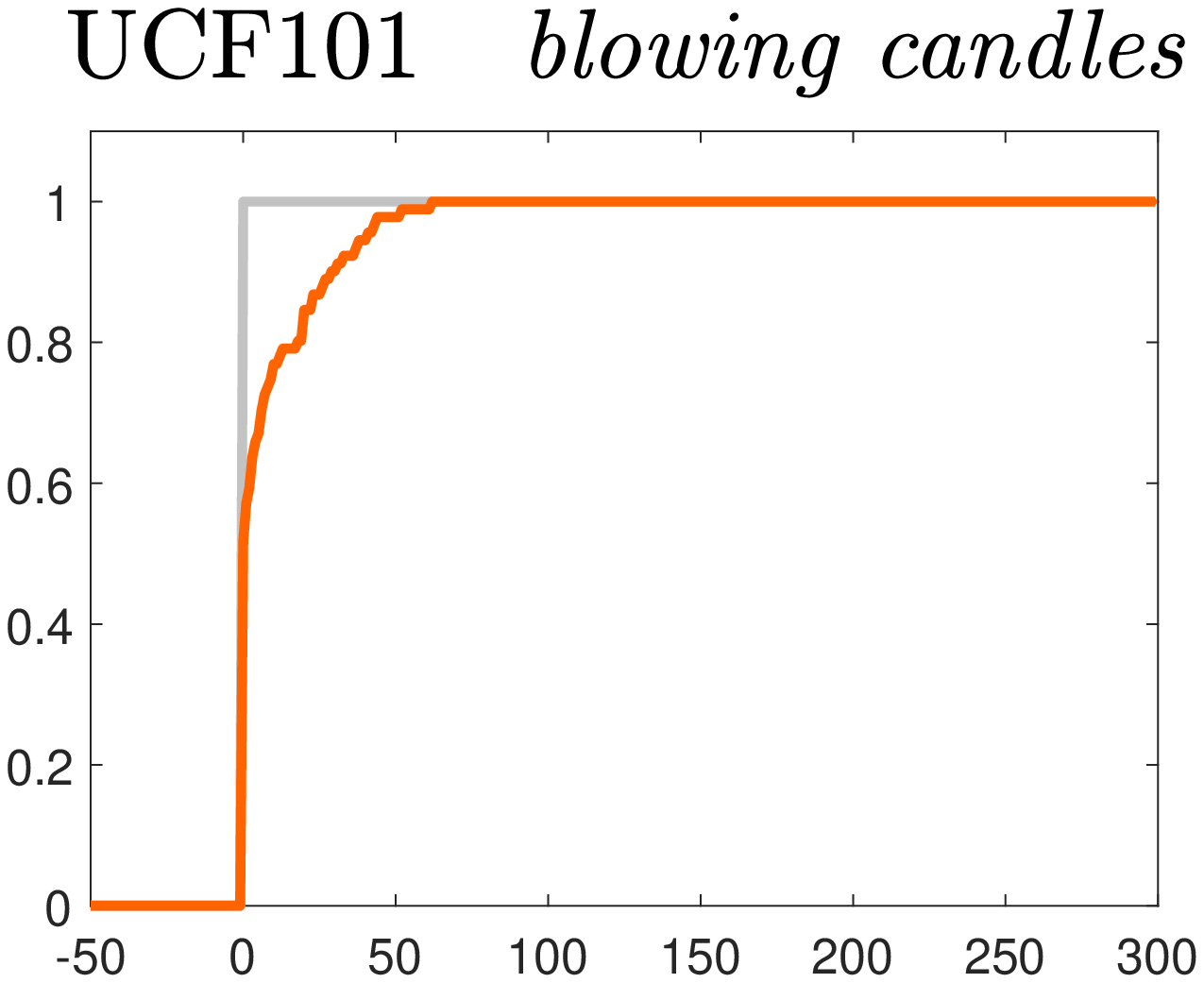}\hspace{-2.2pt}
\includegraphics[width=0.16\textwidth]{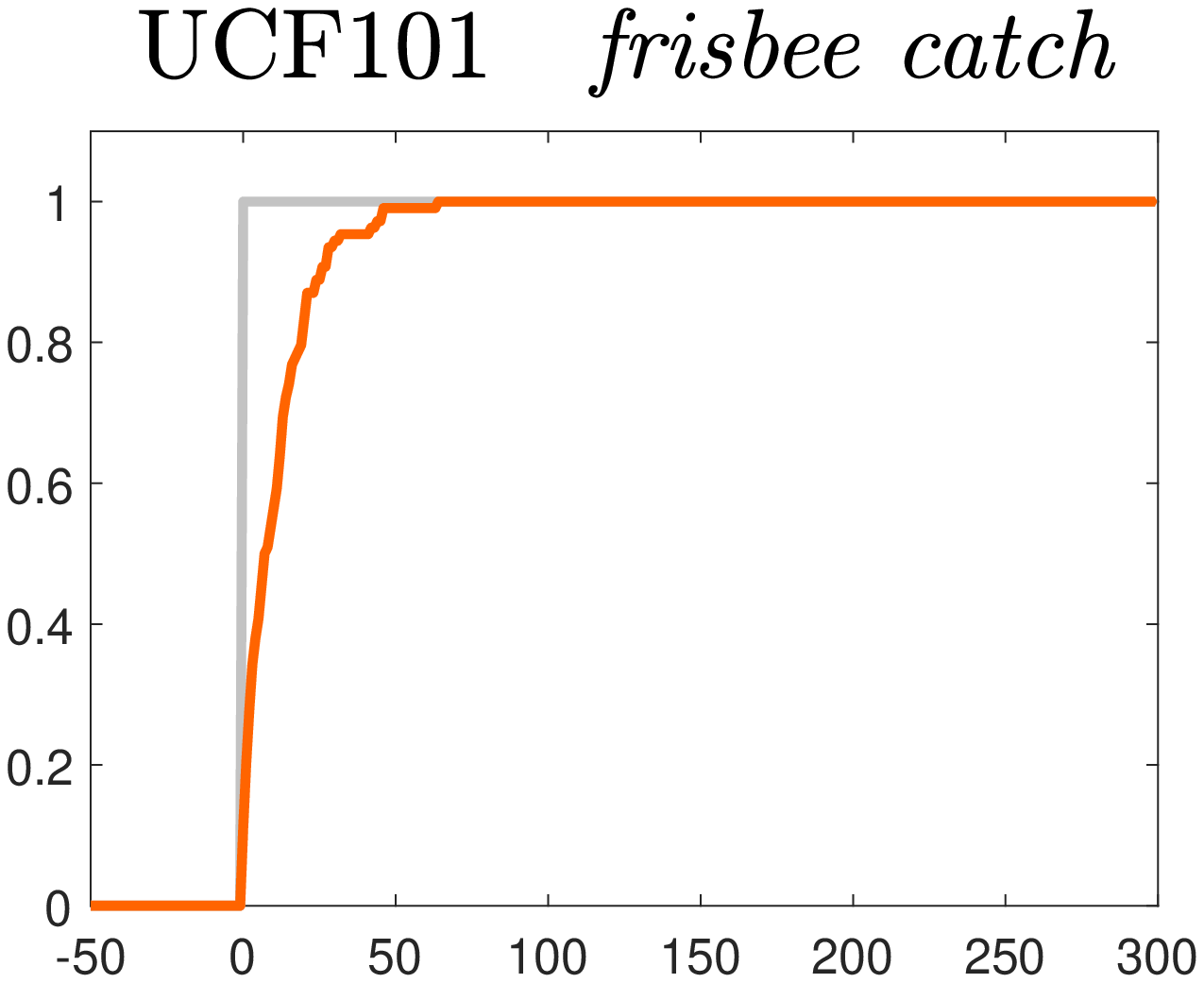}\hspace{-2.2pt}
\includegraphics[width=0.16\textwidth]{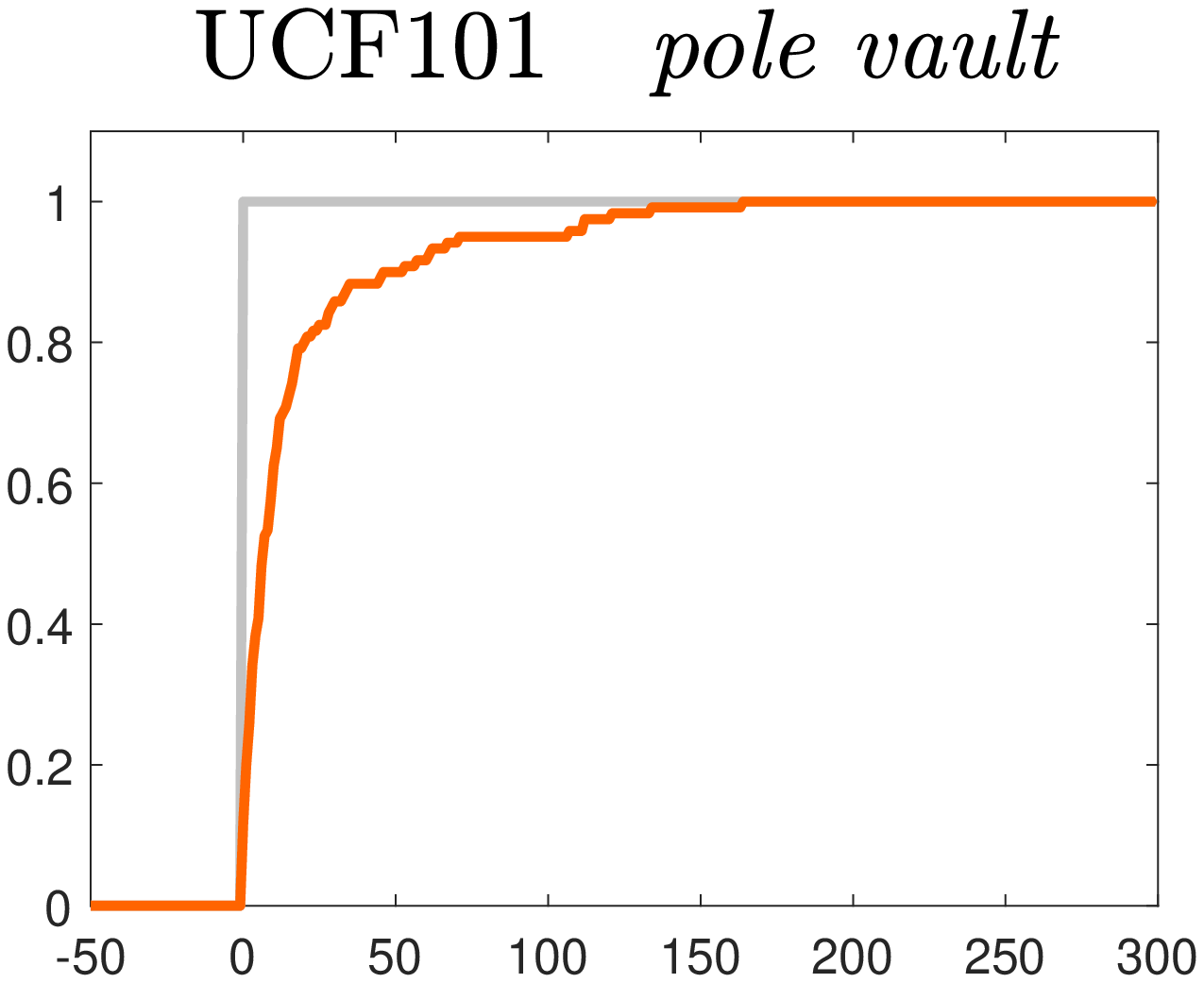}\hspace{-2.2pt}
\includegraphics[width=0.16\textwidth]{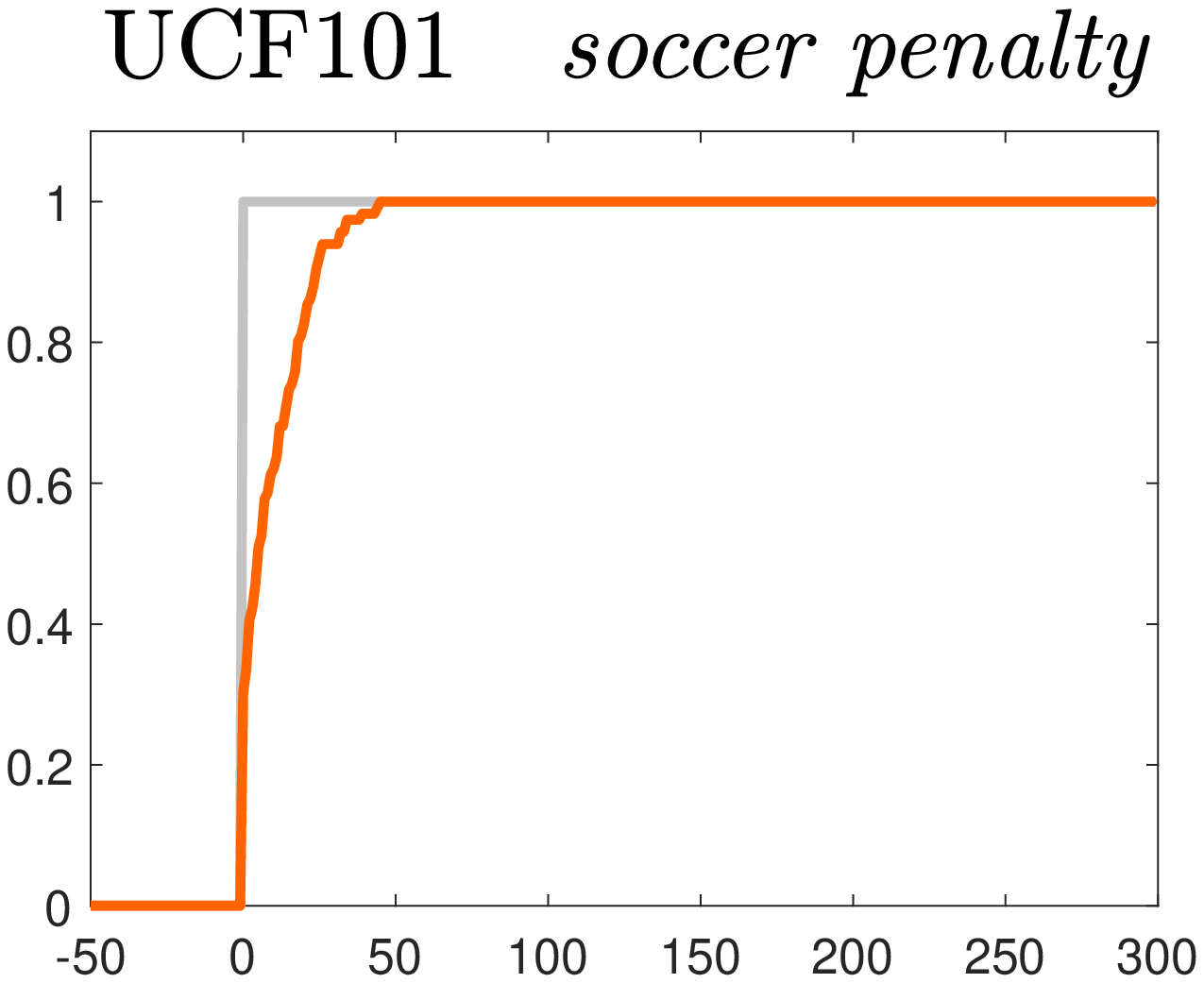}\hspace{-2.2pt}\\
\includegraphics[width=0.16\textwidth]{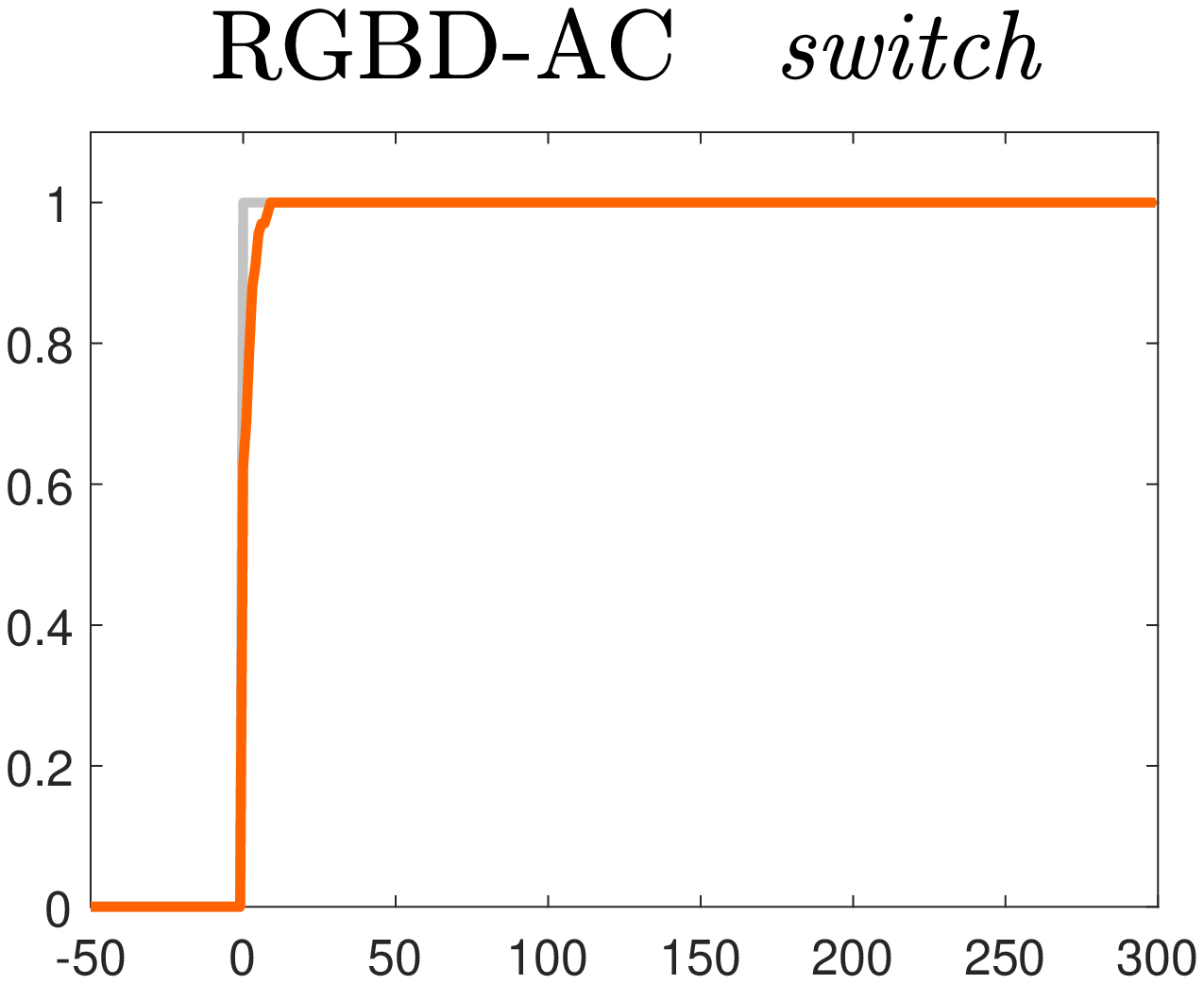}\hspace{-2.2pt}
\includegraphics[width=0.16\textwidth]{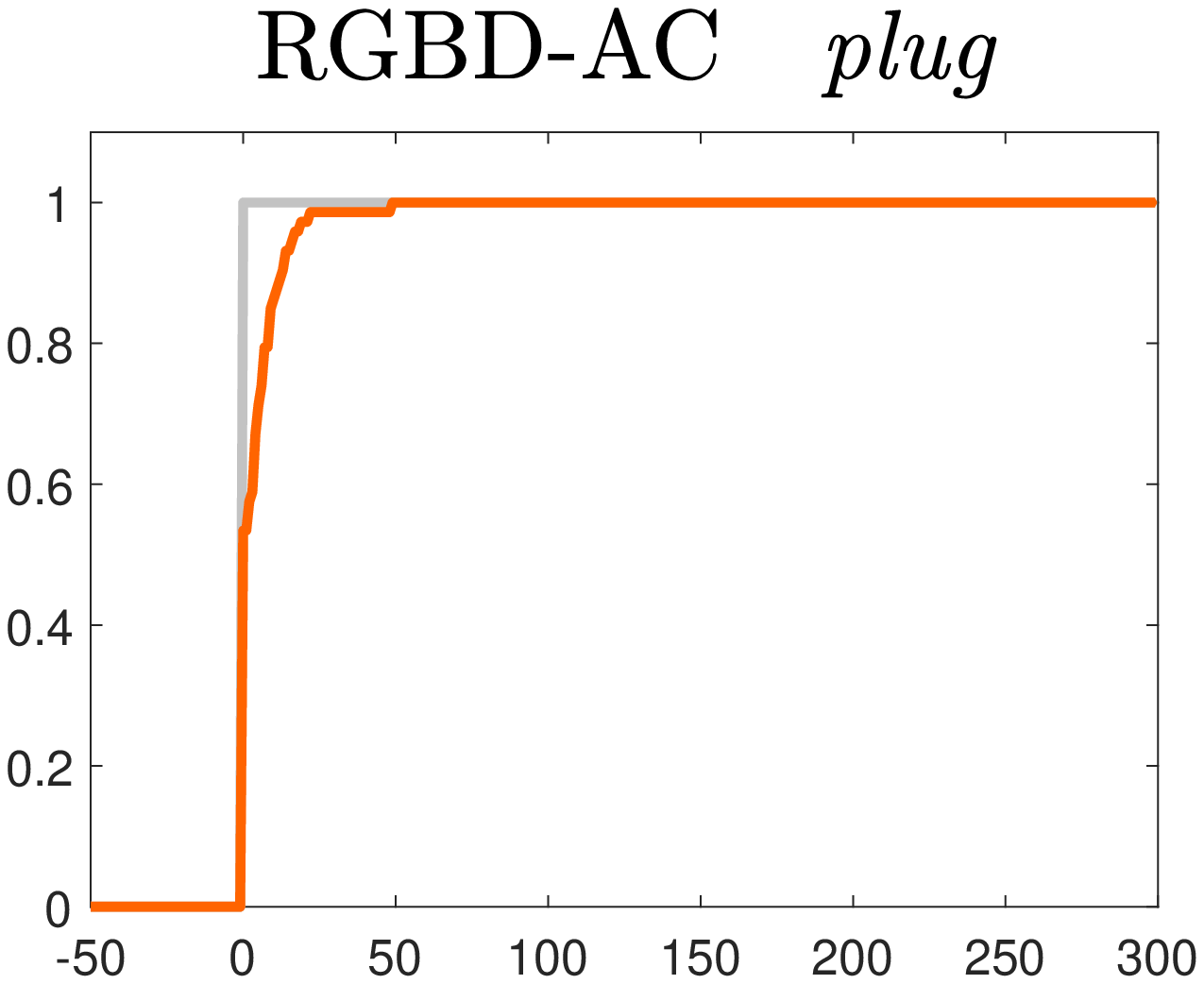}\hspace{-2.2pt}
\includegraphics[width=0.16\textwidth]{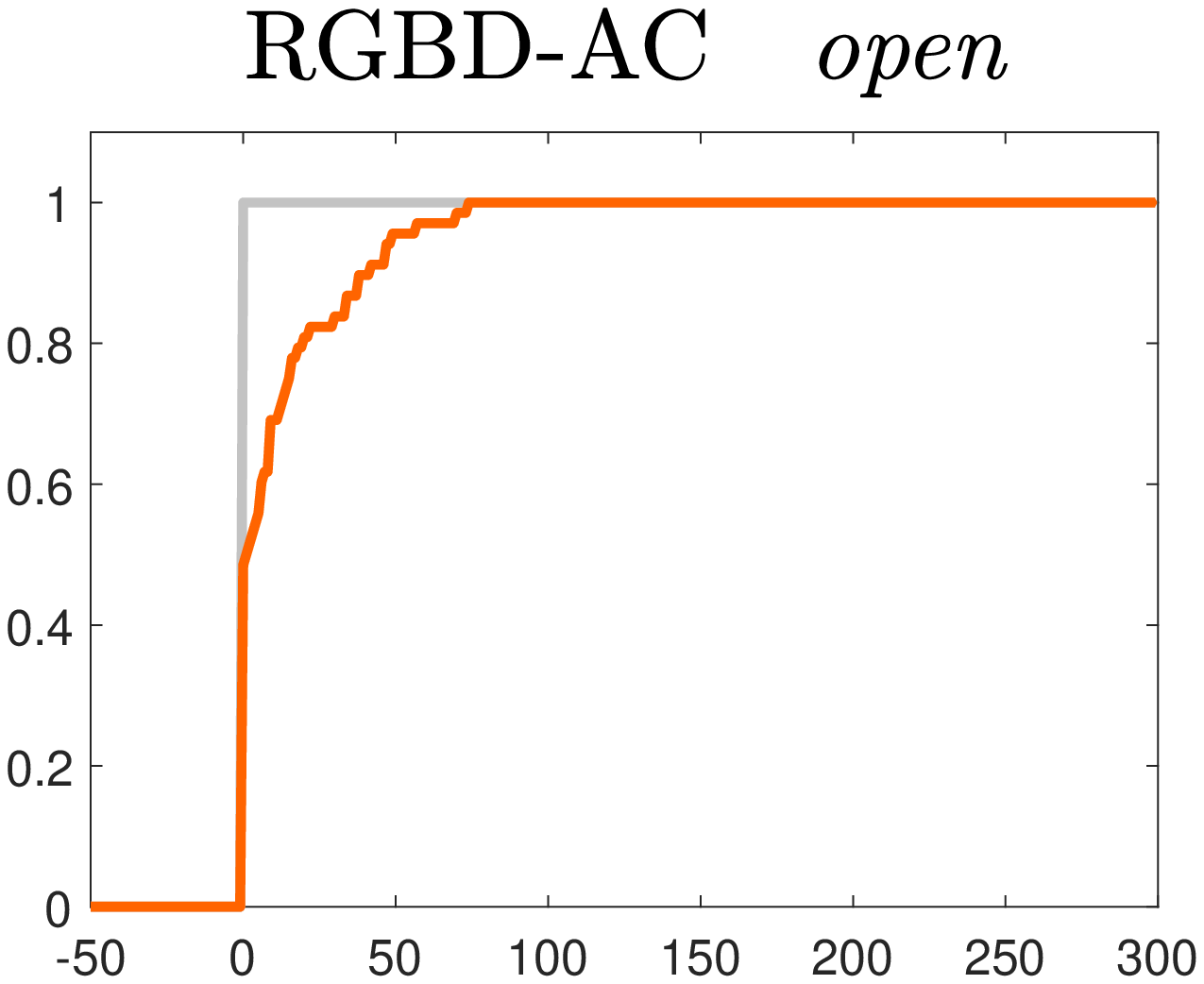}\hspace{-2.2pt}
\vspace{3pt}
\includegraphics[width=0.16\textwidth]{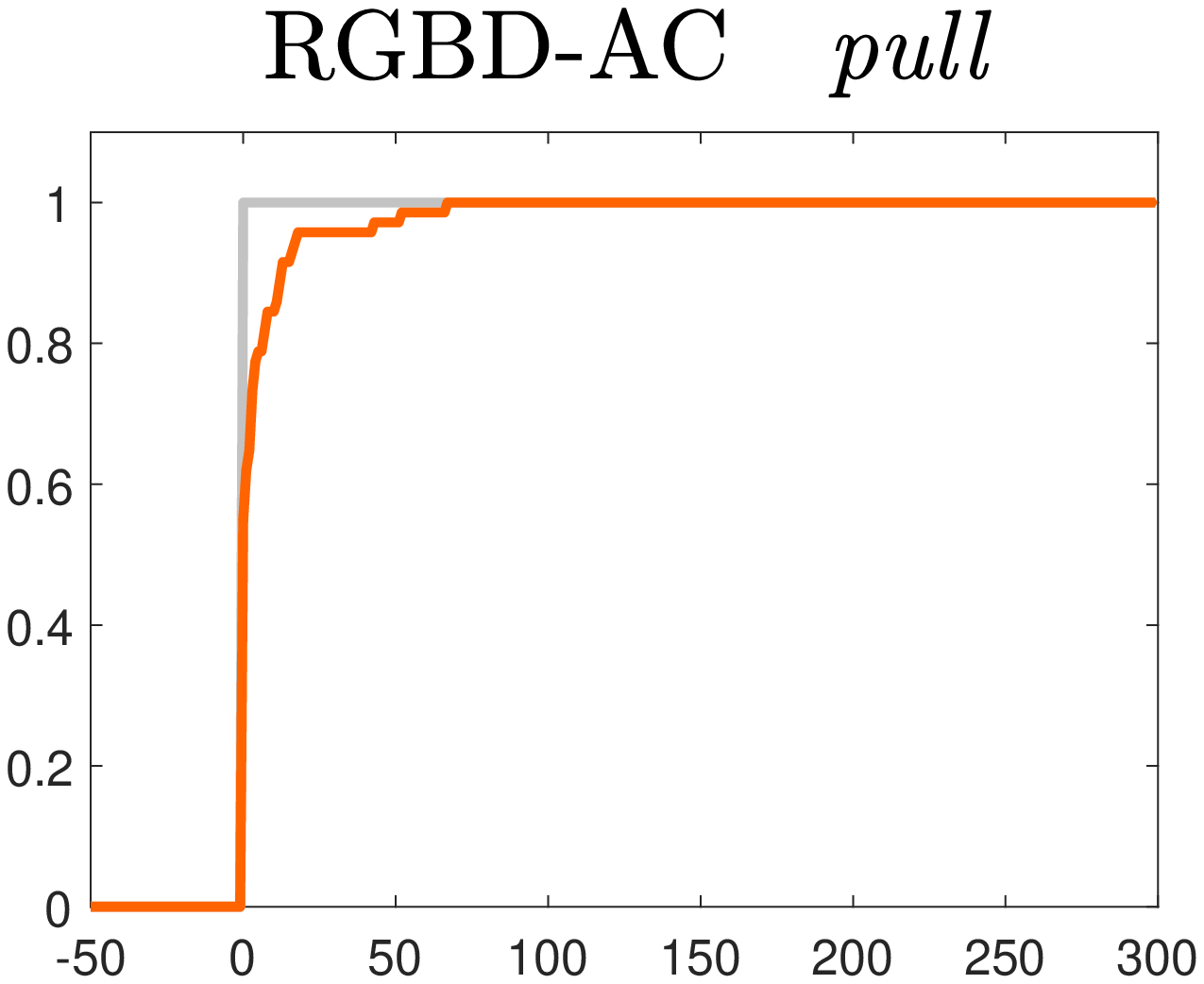}\hspace{-2.2pt}
\includegraphics[width=0.16\textwidth]{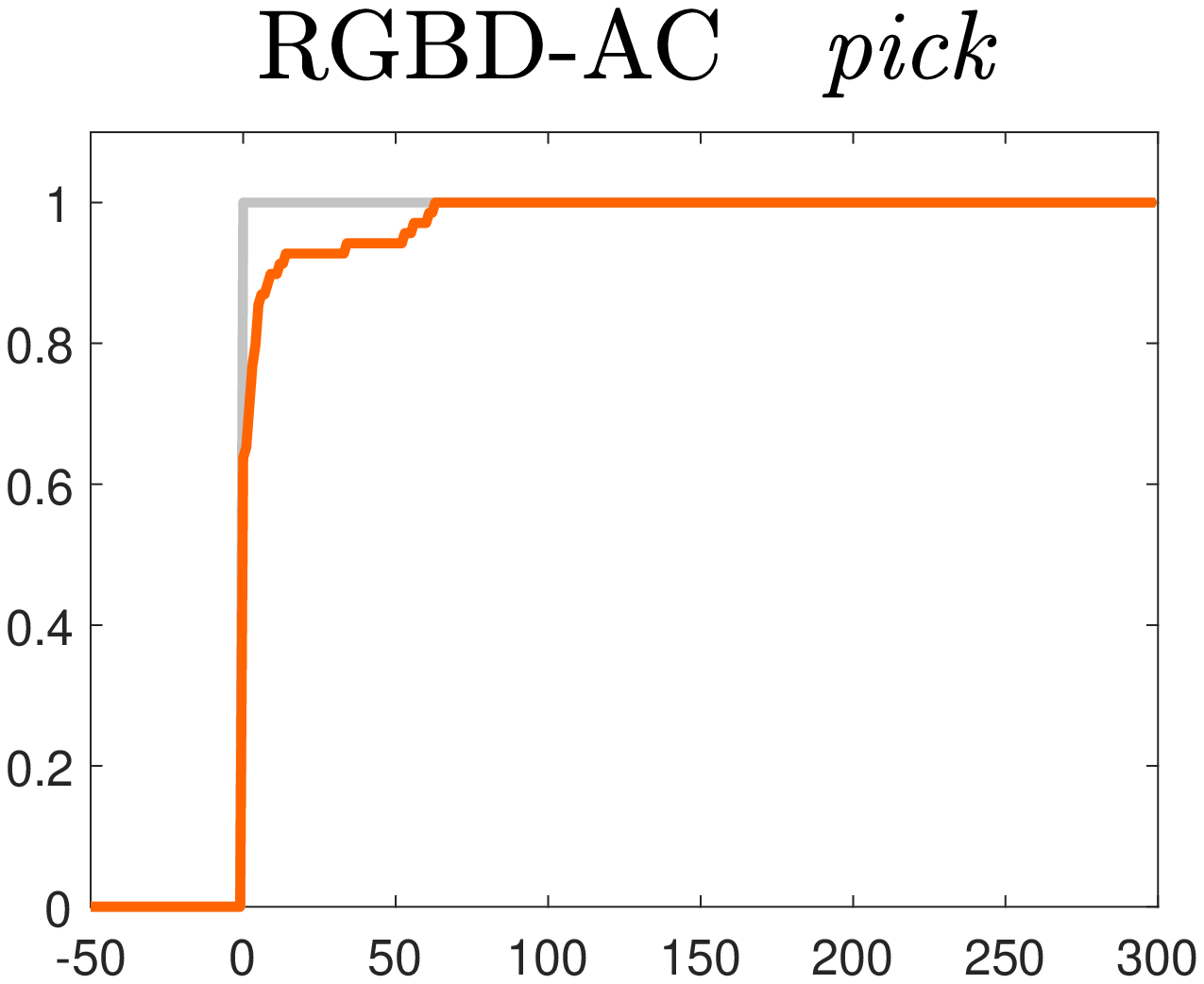}\hspace{-2.2pt}
\includegraphics[width=0.16\textwidth]{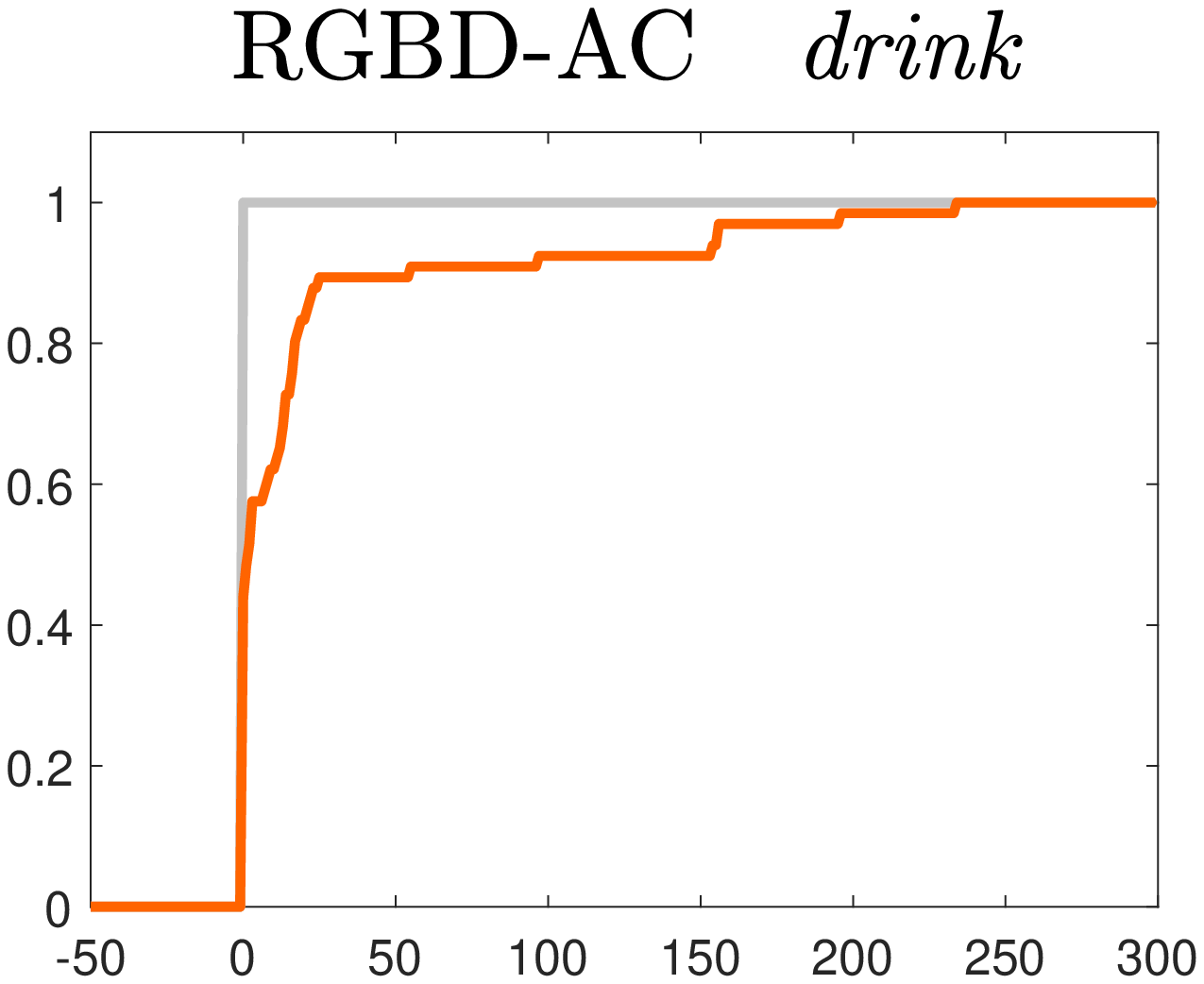}\hspace{-2.2pt}\\

\centering
\includegraphics[width=0.4\textwidth]{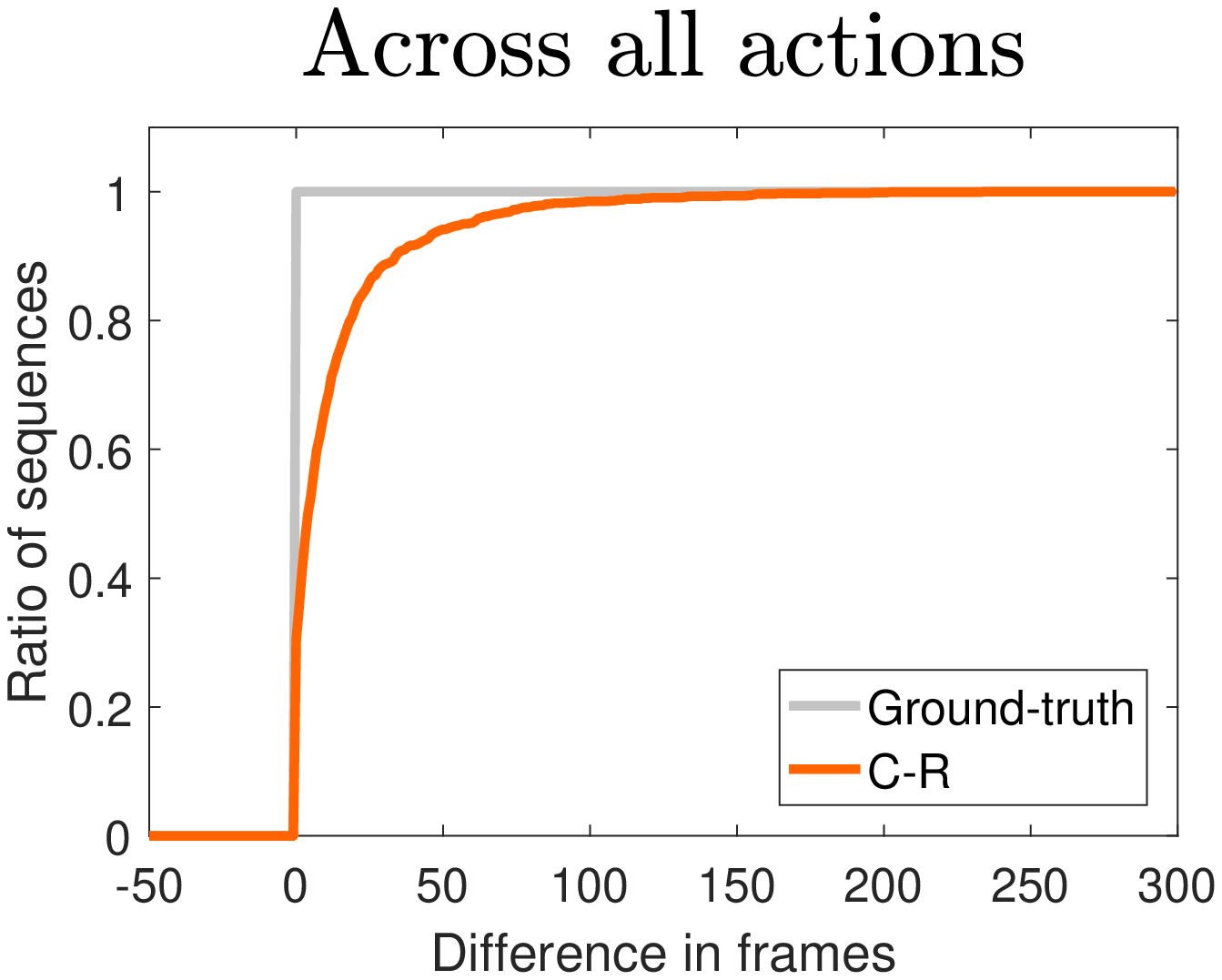}\hspace{-2.2pt}\\
\caption{Cumulative percentage of sequences where the completion moment is detected within $x$ frames. Acceptance threshold $x$ is shown on the x-axis. Results are shown for each of the 16 actions as well as across all actions.} 
\label{fig:res_plot}
\end{figure}

\vspace*{-12pt}
\section{Conclusion and Future Work}
\label{sec:Conc}
\vspace{-4pt}

This paper presents action \textit{completion moment} detection as the task of localising the moment in time when a human observer believes an action's goal has been achieved. 
The approach goes beyond recognition of completion towards a fine-grained perception of completion.  
We use a supervised approach for detecting completion per action, and propose an end-to-end trainable recurrent model. We show that individual frames can contribute to predicting a sequence-level \textit{completion moment} via voting, and propose four methods to accumulate frame-level votes. Results show that using classification voting for pre-completion frames, and regression-voting for post-completion frames achieves the overall best result. 

We foresee the proposed temporal model as a powerful learning method for moment detection in actions, for and beyond action completion.
We aim to pursue two directions for future work. First, we plan to extend our work to untrimmed videos and propose temporal models able to detect multiple \textit{completion moments}. Second, we shall explore {weakly-supervised approaches to \textit{completion moment} detection}.


\begin{appendices}
\section{}

\begin{table} 

\begin{center}
\fontsize{6}{6} \selectfont{
\def\arraystretch{1.3}
\begin{tabular}{|c|c|c|c|c|c|c|c|c|c|}
\cline{5-10}
\multicolumn{4}{c|}{ } & \multicolumn{6}{c|}{Accuracy} \\ \cline{4-10}
\multicolumn{3}{c|}{ } & No. & \textbf{Pre-V} & \textbf{$V_R^T$} & \textbf{C-C} & \textbf{R-R} & \textbf{R-C} & \textbf{C-R} \\ \hline
\multirow{15}{*}{\rotatebox[origin=c]{90}{HMDB}} & \multirow{2}{*}{\textit{catch}} & complete & 99 & 77.3 & 79.1 & 75.9 & 80.5 & 76.7 & 82.3 \\ \cline{3-10}
& & incomplete & 0 & - & - & - & - & - & - \\ \cline{3-10}
& & \textbf{total} & 99 & 77.3 & 79.1 & 75.9 & 80.5 & 76.7 & \textbf{82.3}   \\ \cline{2-10}

&\multirow{2}{*}{\textit{drink}} & complete & 96 & 76.6 & 68.5 & 72.0 & 77.3 & 75.3 & 80.0 \\ \cline{3-10}
& & incomplete & 4 & 92.4 & 87.4 & 99.5 & 94.0 & 88.3 & 91.7 \\ \cline{3-10}
& & \textbf{total} & 100 & 77.3 & 69.3 & 73.2 & 78.0 & 75.9 & \textbf{80.5} \\ \cline{2-10}

&\multirow{2}{*}{\textit{pick}} & complete & 76 & 79.4 & 75.4 & 78.2 & 79.4 & 77.5 & 82.6 \\ \cline{3-10}
& & incomplete & 22 & 84.4 & 92.0 & 84.1 & 81.5 & 66.4 & 88.9 \\ \cline{3-10}
& & \textbf{total} & 98 & 80.6 & 79.5  & 79.7 & 79.9 & 74.7 & \textbf{84.2}   \\ \cline{2-10}

&\multirow{2}{*}{\textit{pour}} & complete & 98 & 77.3 & 68.5 & 71.9 & 80.7 & 79.5 & 81.9 \\ \cline{3-10}
& & incomplete & 1 & 4.5 & 50.5 & 2.7 & 17.1 & 9.0 & 22.5 \\ \cline{3-10}
& & \textbf{total} & 99 & 76.5 & 68.3  & 71.1 & 80.0 & 78.7 & \textbf{81.2} \\ \cline{2-10}

&\multirow{2}{*}{\textit{throw}} & complete & 95 & 67.2 & 73.3 & 61.7 & 74.1 & 64.9 & 79.5 \\ \cline{3-10}
& & incomplete & 3 & 100.0 & 95.6 & 100.0 & 84.6 & 86.0 & 100.0 \\ \cline{3-10}
& & \textbf{total} & 98 & 68.7 & 74.3 & 63.4 & 74.6 & 65.8 & \textbf{80.4} \\ \hline \hline

\multirow{15}{*}{\rotatebox[origin=c]{90}{UCF101}} &\multirow{2}{*}{\textit{basketball}} & complete & 102 & 84.7 & 73.1 & 80.3 & 79.6 & 78.2 & 81.1\\ \cline{3-10}
& & incomplete & 32 & 92.3 & 93.9 & 97.7 & 79.2 & 82.0 & 97.8\\ \cline{3-10}
& & \textbf{total} & 134 & \textbf{86.5} & 78.0 & 84.5 & 79.5 & 79.1 & 85.1 \\ \cline{2-10}

&\multirow{2}{*}{\textit{blowing candles}} & complete & 59 & 80.3 & 80.7 & 78.5 & 78.4 & 67.9 & 84.1 \\ \cline{3-10}
& & incomplete & 50 & 94.2 & 96.8 & 95.3 & 90.6 & 89.7 & 98.5\\ \cline{3-10}
& & \textbf{total} & 109 & 86.8 & 88.3 & 86.4 & 84.2 & 78.2 & \textbf{90.9} \\ \cline{2-10}

&\multirow{2}{*}{\textit{frisbee catch}} & complete & 125 & 81.7 & 84.1 & 80.3 & 78.3 & 74.6 & 85.9 \\ \cline{3-10}
& & incomplete & 0 & - & - & - & - & - & -\\ \cline{3-10}
& & \textbf{total} & 125 & 81.7 & 84.1  & 80.3 & 78.3 & 74.6 & \textbf{85.9}  \\ \cline{2-10}

&\multirow{2}{*}{\textit{pole vault}} & complete & 142 & 85.0 & 83.3 & 82.4 & 88.5 & 79.8 & 90.6 \\ \cline{3-10}
& & incomplete & 3 & 87.4 & 81.8 & 88.5 & 84.0 & 92.1 & 90.9 \\ \cline{3-10}
& & \textbf{total} & 145 & 85.0 & 83.3 & 82.6 & 88.4 & 80.1 & \textbf{90.6} \\ \cline{2-10}

&\multirow{2}{*}{\textit{soccer penalty}} & complete & 95 & 85.3 & 83.5 & 84.4 & 86.8 & 83.6 & 86.9 \\ \cline{3-10}
& & incomplete & 42 & 86.0 & 93.5 & 88.8 & 87.7 & 90.0 & 92.1 \\ \cline{3-10}
& & \textbf{total} & 137 & 85.5 & 86.6 & 85.8 & 87.1 & 85.6 & \textbf{88.5}  \\ \hline \hline

\multirow{18}{*}{\rotatebox[origin=c]{90}{RGBD-AC}} &\multirow{2}{*}{\textit{switch}} & complete & 35 & 99.8 & 88.7 & 99.8 & 96.3 & 86.0 & 98.0\\ \cline{3-10}
& & incomplete & 32 & 100 & 99.7 & 100.0 & 100.0 & 100.0 & 100.0\\ \cline{3-10}
& & \textbf{total} & 67 & \textbf{99.9} & 93.9 & \textbf{99.9} & 98.1 & 92.7 & 98.9  \\ \cline{2-10}

&\multirow{2}{*}{\textit{plug}} & complete & 37 & 96.8 & 90.0 & 97.1 & 92.8 & 86.3 &  94.4\\ \cline{3-10}
& & incomplete & 36 & 99.8 & 96.4 & 100.0 & 99.4 & 100.0 & 100.0\\ \cline{3-10}
& & \textbf{total} & 73 & 98.3 & 93.2 & \textbf{98.5} & 96.1 & 93.0 & 97.2  \\ \cline{2-10}

&\multirow{2}{*}{\textit{open}} & complete & 36 & 84.6 & 75.3 & 83.1 & 86.9 & 80.3 & 80.9\\ \cline{3-10}
& & incomplete & 32 & 98.3 & 98.4 & 100.0 & 86.4 & 80.5 & 100.0\\ \cline{3-10}
& & \textbf{total} & 68 & \textbf{91.1} & 86.1 & \textbf{91.1} & 86.7 & 80.4 & 89.9  \\ \cline{2-10}

&\multirow{2}{*}{\textit{pull}} & complete & 34 & 96.4 & 85.2 & 95.4 & 95.4 & 85.9 & 95.9\\ \cline{3-10}
& & incomplete & 37 & 98.9 & 92.6 & 100.0 & 92.8 & 96.7 & 98.1\\ \cline{3-10}
& & \textbf{total} & 71 & 97.7 & 89.1 & \textbf{97.8} & 94.1 & 91.5 & 97.0  \\ \cline{2-10}

&\multirow{2}{*}{\textit{pick}} & complete & 33 & 92.4 & 83.3 & 90.9 & 93.0 & 76.3 & 95.4\\ \cline{3-10}
& & incomplete & 36 & 90.7 & 94.3 & 89.0 & 93.4 & 90.2 & 94.5\\ \cline{3-10}
& & \textbf{total} & 69 & 91.5 & 89.1 & 89.9 & 93.2 & 83.6 & \textbf{95.0}  \\ \cline{2-10}

&\multirow{2}{*}{\textit{drink}} & complete & 34 & 89.3 & 66.3 & 83.1 & 92.7 & 87.9 & 92.8\\ \cline{3-10}
& & incomplete & 32 & 87.9 & 92.5 & 87.6 & 89.0 & 83.5 & 91.3\\ \cline{3-10}
& & \textbf{total} & 66 & 88.6 & 79.0 & 85.3 & 90.9 & 85.8 & \textbf{92.1} \\ \hline \hline

\multicolumn{3}{|c|}{\textbf{complete}} & 1196 & 82.3 & 78.1 & 79.6 & 83.1 & 77.7 & 85.6 \\ \hline
\multicolumn{3}{|c|}{\textbf{incomplete}} & 362 & 93.4 & 94.8 & 94.3 & 90.4 & 88.8 & 96.1 \\ \hline
\multicolumn{3}{|c|}{\textbf{total}} & 1558 & 85.0 & 82.2 & 83.2 & 84.9 & 80.4 & \textbf{88.1} \\ \hline

\end{tabular}}
\end{center}
\caption{Results on 16 actions, comparing frame-level classification, last-frame regression and the four sequence-level voting schemes.}
\label{tab:acc}
\end{table}


\begin{table} 

\begin{center}
\fontsize{6}{6} \selectfont{
\def\arraystretch{1.3}
\begin{tabular}{|c|c|c|c|c|c|c|c|c|}
\cline{4-9}
\multicolumn{3}{c|}{ } & \multicolumn{6}{c|}{$RD$} \\ \cline{4-9}
\multicolumn{3}{c|}{ } & \textbf{Pre-V} & \textbf{$V_R^T$} & \textbf{C-C} & \textbf{R-R} & \textbf{R-C} & \textbf{C-R} \\ \hline
\multirow{15}{*}{\rotatebox[origin=c]{90}{HMDB}} & \multirow{3}{*}{\textit{catch}} & complete & 0.23 & 0.21 & 0.24 & 0.20 & 0.23 &\textbf{0.18}  \\ \cline{3-9}
& & incomplete & - & - & - & - & - & - \\ \cline{3-9}
& & \textbf{total} & 0.23 & 0.21 & 0.24 & 0.20 & 0.23 &\textbf{0.18} \\ \cline{2-9}

&\multirow{3}{*}{\textit{drink}} & complete & 0.21 & 0.32 & 0.28 & 0.23 & 0.25 & 0.20 \\ \cline{3-9}
& & incomplete & 0.38 & 0.13 & 0.00 & 0.06 & 0.12 & 0.08 \\ \cline{3-9}
& & \textbf{total} & 0.21 & 0.31 & 0.27 & 0.22 & 0.24 & \textbf{0.19} \\ \cline{2-9}

&\multirow{3}{*}{\textit{pick}} & complete & 0.20 & 0.25 & 0.22 & 0.21 & 0.23 & 0.17 \\ \cline{3-9}
& & incomplete & 0.29 & 0.08 & 0.16 & 0.18 & 0.34 & 0.11 \\ \cline{3-9}
& & \textbf{total} & 0.22 & 0.20 & 0.20 & 0.20 & 0.25 & \textbf{0.16} \\ \cline{2-9}

&\multirow{3}{*}{\textit{pour}} & complete & 0.22 & 0.31 & 0.28 & 0.19 & 0.20 & 0.18 \\ \cline{3-9}
& & incomplete & 0.97 & 0.50 & 0.97 & 0.83 & 0.91 & 0.77 \\ \cline{3-9}
& & \textbf{total} & 0.23 & 0.32 & 0.29 & 0.20 & 0.21 & \textbf{0.19} \\ \cline{2-9}

&\multirow{3}{*}{\textit{throw}} & complete & 0.33 & 0.27 & 0.38 & 0.26 & 0.35 & 0.21 \\ \cline{3-9}
& & incomplete & 0.00 & 0.04 & 0.00 & 0.15 & 0.14 & 0.00 \\ \cline{3-9}
& & \textbf{total} & 0.32 & 0.26 & 0.37 & 0.25 & 0.34 & \textbf{0.20} \\ \hline \hline

\multirow{15}{*}{\rotatebox[origin=c]{90}{UCF101}} &\multirow{3}{*}{\textit{basketball}} & complete & 0.19 & 0.27 & 0.20 & 0.20 & 0.22 & 0.19\\ \cline{3-9}
& & incomplete & 0.27 & 0.06 & 0.02 & 0.21 & 0.18 & 0.02 \\ \cline{3-9}
& & \textbf{total} & 0.21 & 0.22 & 0.16 & 0.20 & 0.21 & \textbf{0.15} \\ \cline{2-9}

&\multirow{3}{*}{\textit{blowing candles}} & complete & 0.20 & 0.19 & 0.22 & 0.22 & 0.32 & 0.16 \\ \cline{3-9}
& & incomplete & 0.11 & 0.03 & 0.05 & 0.09 & 0.10 & 0.02 \\ \cline{3-9}
& & \textbf{total} & 0.16 & 0.12 & 0.14 & 0.16 & 0.22 & \textbf{0.09} \\ \cline{2-9}

&\multirow{3}{*}{\textit{frisbee catch}} & complete & 0.24 & 0.16 & 0.20 & 0.22 & 0.25 & 0.14 \\ \cline{3-9}
& & incomplete & - & - & - & - & - & -\\ \cline{3-9}
& & \textbf{total} & 0.24 & 0.16 & 0.20 & 0.22 & 0.25 & \textbf{0.14} \\ \cline{2-9}

&\multirow{3}{*}{\textit{pole vault}} & complete & 0.19 & 0.17 & 0.18 & 0.12 & 0.20 & 0.09\\ \cline{3-9}
&  & incomplete & 0.18 & 0.18 & 0.11 & 0.16 & 0.08 & 0.09 \\ \cline{3-9}
& & \textbf{total} & 0.19 & 0.17 & 0.17 & 0.12 & 0.20 & \textbf{0.09} \\ \cline{2-9}

&\multirow{3}{*}{\textit{soccer penalty}} & complete & 0.15 & 0.17 & 0.16 & 0.13 & 0.16 & 0.13\\ \cline{3-9}
& & incomplete & 0.16 & 0.06 & 0.11 & 0.12 & 0.10 & 0.08 \\ \cline{3-9}
& & \textbf{total} & 0.15 & 0.13 & 0.14 & 0.13 & 0.14 & \textbf{0.11} \\ \hline \hline

\multirow{18}{*}{\rotatebox[origin=c]{90}{RGBD-AC}} &\multirow{3}{*}{\textit{switch}} & complete & 0.00 & 0.11 & 0.00 & 0.04 & 0.14 & 0.02\\ \cline{3-9}
& & incomplete & 0.00 & 0.00 & 0.00 & 0.00 & 0.00 & 0.00 \\ \cline{3-9}
& & \textbf{total} & \textbf{0.00} & 0.06 & \textbf{0.00} & 0.02 & 0.07 & 0.01 \\ \cline{2-9}

&\multirow{3}{*}{\textit{plug}} & complete & 0.04 & 0.10 & 0.03 & 0.07 & 0.14 & 0.06 \\ \cline{3-9}
& & incomplete & 0.01 & 0.04 & 0.00 & 0.01 & 0.00 & 0.00 \\ \cline{3-9}
& & \textbf{total} & 0.02 & 0.07 & \textbf{0.01} & 0.04 & 0.07 & 0.03 \\ \cline{2-9}

&\multirow{3}{*}{\textit{open}} & complete & 0.13 & 0.25 & 0.17 & 0.13 & 0.20 & 0.19 \\ \cline{3-9}
& & incomplete & 0.12 & 0.02 & 0.00 & 0.14 & 0.19 & 0.00 \\ \cline{3-9}
& & \textbf{total} & 0.12 & 0.14 & \textbf{0.09} & 0.13 & 0.20 & 0.10 \\ \cline{2-9}

&\multirow{3}{*}{\textit{pull}} & complete & 0.05 & 0.15 & 0.05 & 0.05 & 0.14 & 0.04 \\ \cline{3-9}
& & incomplete & 0.14 & 0.07 & 0.00 & 0.07 & 0.03 & 0.02 \\ \cline{3-9}
& & \textbf{total} & 0.10 & 0.11 & \textbf{0.02} & 0.06 & 0.08 & 0.03 \\ \cline{2-9}

&\multirow{3}{*}{\textit{pick}} & complete & 0.09 & 0.17 & 0.09 & 0.07 & 0.24 & 0.05 \\ \cline{3-9}
& & incomplete & 0.13 & 0.06 & 0.11 & 0.07 & 0.10 & 0.05\\ \cline{3-9}
& & \textbf{total} & 0.11 & 0.11 & 0.10 & 0.07 & 0.16 & \textbf{0.05} \\ \cline{2-9}

&\multirow{3}{*}{\textit{drink}} & complete & 0.09 & 0.34 & 0.17 & 0.07 & 0.12 & 0.07 \\ \cline{3-9}
& & incomplete & 0.12 & 0.08 & 0.12 & 0.11 & 0.17 & 0.09\\ \cline{3-9}
& & \textbf{total}  & 0.11 & 0.21 & 0.15 & 0.09 & 0.14 & \textbf{0.08} \\ \hline \hline

\multicolumn{3}{|c|}{\textbf{complete}} & 0.19 & 0.22 & 0.20 & 0.17 & 0.22 & 0.14 \\ \hline
\multicolumn{3}{|c|}{\textbf{incomplete}} & 0.13 & 0.05 & 0.06 & 0.10 & 0.11 & 0.04 \\ \hline
\multicolumn{3}{|c|}{\textbf{total}} & 0.17 & 0.18 & 0.17 & 0.15 & 0.20 & \textbf{0.12} \\ \hline

\end{tabular}}
\end{center}
\caption{Results on 16 actions, comparing frame-level classification, last-frame regression and the four sequence-level voting schemes.}
\label{tab:rd}
\end{table}

For completion, we present the full set of results in two tables. 

\begin{itemize}

\item Table \ref{tab:acc} presents the accuracy for complete and incomplete sequences of the three datasets separately. For the 362 incomplete sequences, across all datasets, the accuracy when using the C-R method is 96.1\%. For the 1196 complete sequences, the accuracy when using the C-R method is 85.6\%. 

\item Table \ref{tab:rd} shows the RD evaluation measure for the complete and incomplete sequences of the three datasets separately. Again, C-R voting has the lowest RD error with 0.14 for all complete sequences and 0.04 for all incomplete sequences.


\end{itemize}

\clearpage

\end{appendices}

\bibliography{egbib}
\end{document}